\documentclass[lettersize,journal]{IEEEtran}
\usepackage{amsmath,amsfonts}
\usepackage{array}
\usepackage[caption=false,font=normalsize,labelfont=sf,textfont=sf]{subfig}
\usepackage{textcomp}
\usepackage{stfloats}
\usepackage{url}
\usepackage{verbatim}
\usepackage{graphicx}
\usepackage{times}
\usepackage{epsfig}
\usepackage{amsmath}
\usepackage{amssymb}
\usepackage{float}
\usepackage{bbding}
\usepackage{amsmath}
\usepackage{amssymb}
\usepackage{booktabs}
\usepackage{soul}
\usepackage{color}
\usepackage{colortbl}
\definecolor{Gray}{gray}{0.9}
\definecolor{baselinecolor}{gray}{0.9}
\usepackage{pifont} 
\usepackage{multirow}
\usepackage{overpic}
\usepackage{textpos}
\usepackage{array}
\usepackage{makecell}
\usepackage{footnote}
\newcommand{\tablestyle}[2]{\setlength{\tabcolsep}{#1}\renewcommand{\arraystretch}{#2}\centering\footnotesize}
\usepackage{utfsym}
\usepackage{threeparttable}
\usepackage{algpseudocode} 
\usepackage{bm}
\usepackage[normalem]{ulem}
\usepackage{diagbox}
\usepackage{multirow}
\usepackage{cellspace}

\usepackage{amsmath,amsfonts}
\usepackage[caption=false,font=normalsize,labelfont=sf,textfont=sf]{subfig}
\usepackage{algorithmicx,algorithm} 

\newcommand{\ym}[1]{{\color{black}#1}}

\def\STATE{\State}
\def\STATE{\State}
\def\FOR{\For}
\def\WHILE{\While}

\usepackage[pagebackref=true,breaklinks=true,letterpaper=true,colorlinks,bookmarks=false]{hyperref}
\usepackage[capitalize]{cleveref}
\crefname{section}{Sec.}{Secs.}
\crefname{section}{Section}{Sections}
\crefname{table}{Table}{Tables}
\crefname{table}{Tab.}{Tabs.}

\definecolor{yellow}{RGB}{255,239,213}

\def\BibTeX{{\rm B\kern-.05em{\sc i\kern-.025em b}\kern-.08em
    T\kern-.1667em\lower.7ex\hbox{E}\kern-.125emX}}
\usepackage{balance}
\begin{document}
\title{Continual Action Assessment via Task-Consistent Score-Discriminative Feature Distribution Modeling}
\author{Yuan-Ming Li, Ling-An Zeng, Jing-Ke Meng and Wei-Shi Zheng 
\thanks{
This work was supported partially by the NSFC(U21A20471, U1911401, 62206315), Guangdong NSF Project (No. 2023B1515040025, 2020B1515120085), and Guangzhou Basic and Applied Basic Research Scheme(2024A04J4067).

Yuan-Ming Li and Jing-Ke Meng are with the School of Computer Science and Engineering, Sun Yat-sen University, Guangzhou 510006, China. Ling-An Zeng is with the School of Artificial Intelligence, Sun Yat-sen University, Zhuhai 519082, China. 

Wei-Shi Zheng is with the School of Computer Science and EngineeringSun Yat-sen University, Guangzhou, Guangdong 510275, China, also with theGuangdong Key Laboratory of Information Security Technology, Sun Yat-senUniversity, Guangzhou, Guangdong 510275, China, and also with the KeyLaboratory of Machine Intelligence and Advanced Computing, Sun Yat-senUniversity, Ministry of Education, Guangzhou, Guangdong 510275, China.

E-mails: \{liym266, zenglan3\}@mail2.sysu.edu.cn, mengjke@gmail.com, wszheng@ieee.org /zhwshi@mail.sysu.edu.cn. (Corresponding author: Jing-Ke Meng)}}
\markboth{}%
{How to Use the IEEEtran \LaTeX \ Templates}

\maketitle

\begin{abstract}
Action Quality Assessment (AQA) is a task that tries to answer how well an action is carried out. While remarkable progress has been achieved, existing works on AQA assume that all the training data are visible for training at one time, but do not enable continual learning on assessing new technical actions. In this work, we address such a Continual Learning problem in AQA (Continual-AQA), which urges a unified model to learn AQA tasks sequentially without forgetting. 
Our idea for modeling Continual-AQA is to sequentially learn a task-consistent score-discriminative feature distribution,  in which the latent features express a strong correlation with the score labels regardless of the task or action types.
From this perspective, we aim to mitigate the forgetting in Continual-AQA from two aspects.
Firstly, to fuse the features of new and previous data into a score-discriminative distribution, a novel Feature-Score Correlation-Aware Rehearsal is proposed to store and reuse data from previous tasks with limited memory size.
Secondly, an Action General-Specific Graph is developed to learn and decouple the action-general and action-specific knowledge so that the task-consistent score-discriminative features can be better extracted across various tasks. Extensive experiments are conducted to evaluate the contributions of proposed components. The comparisons with the existing continual learning methods additionally verify the effectiveness and versatility of our approach. Data and code are available at https://github.com/iSEE-Laboratory/Continual-AQA.
\end{abstract}

\begin{IEEEkeywords}
Action Quality Assessment, Continual Learning.
\end{IEEEkeywords}

\section{Introduction}
\label{sec:intro}

\IEEEPARstart{A}{ction} Quality Assessment (AQA) aims to assess the quality of action in videos.
Rather than classifying the video samples by their action types \cite{i3d,slowfast,timesformer, SSRL-csvt, Transductive-csvt}, a common setting of AQA is to predict a subtle action quality score for a video action instance. 
Recently, AQA has attached more and more attention for its great potential value for sports skills analysis \cite{JRG,finediving,xu2022likert,zeng2020hybrid}, surgical maneuver training \cite{visa,vtpe,malpani2014pairwise} and so on\cite{whos-better, best}.

Although remarkable progress has been achieved, a remaining challenge in AQA has yet to be explored.
In practical application, {the increasing demand drives the model to sequentially expand its applicability on different tasks (e.g., new technical actions or various daily actions)}. 
A more practical example is that, considering someday you may hope your AI cooking assistant can guide you \ym{in preparing} a French dish, but currently, it may only assess how well you cook Japanese cuisine. 
However, \ym{current AQA methods assume that all the training data is visible for training at one time, 
neglecting the necessity for continual learning to meet evolving demands.
Consequently,} the already trained model may not work on the newly come tasks, resulting in misjudging or misguiding.

\begin{figure}[t]
  \centering
   \includegraphics[width=1\linewidth]{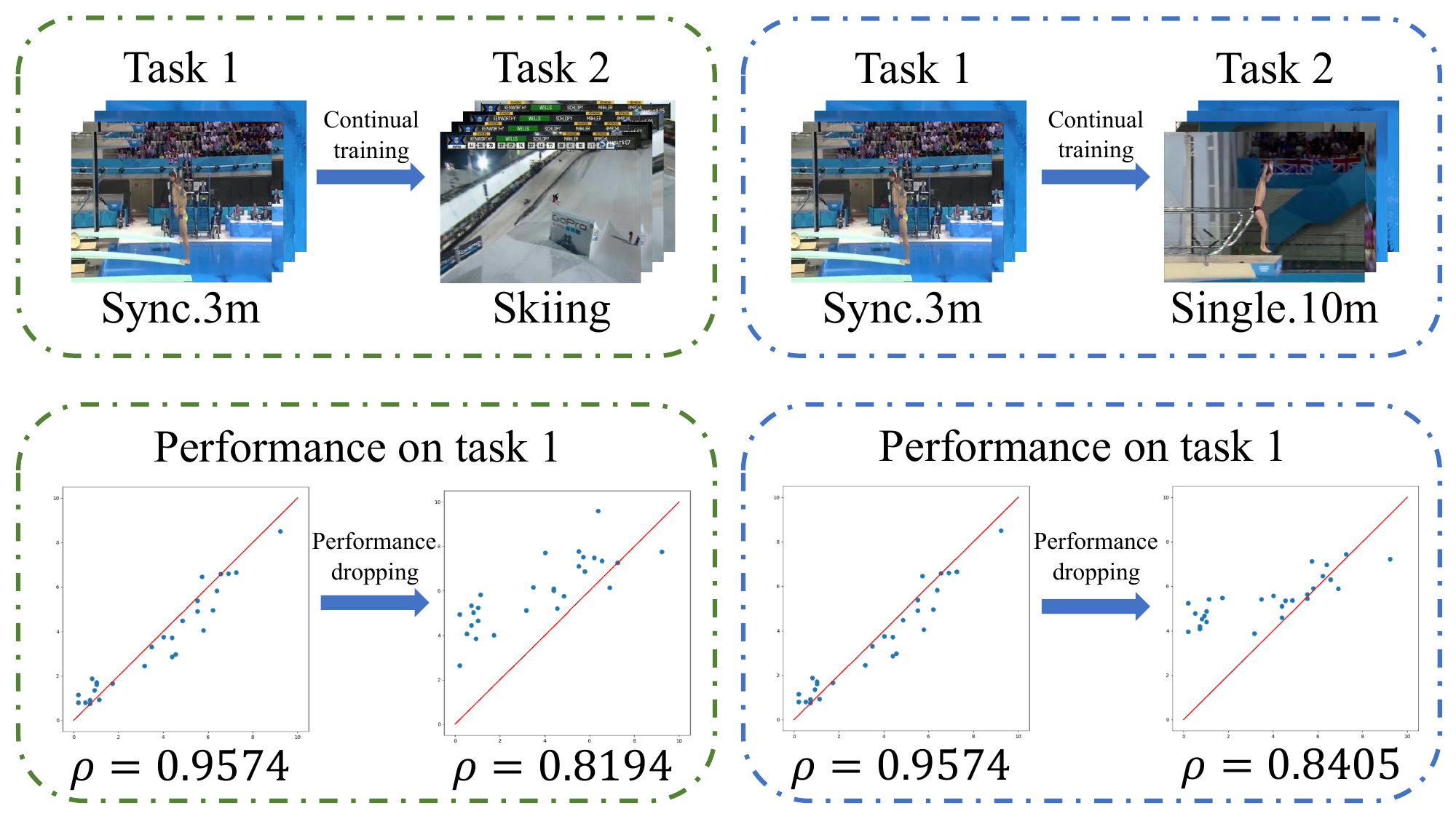}
   \caption{\textbf{Catastrophic forgetting in Continual-AQA.} We sequentially train an AQA model on two tasks and show the performance on the first task in scatter plot and Spearman’s rank correlation coefficient ($\rho$). 
   A higher $\rho$ (i.e., more points close to the red lines) indicates better performance.
   }
   \label{fig:CF_IAQA}
\end{figure}

In this work, we \ym{delve into} a more complex but real scenario, Continual Learning (CL) in AQA, namely Continual-AQA. \ym{Similar to} the other deep neural networks, AQA models also suffer from Catastrophic Forgetting\cite{french1999catastrophic,mccloskey1989catastrophic,goodfellow2013catastrophic}.
As shown in Figure~\ref{fig:CF_IAQA}, when directly fine-tuning an already trained model from one AQA task to another, \ym{there is a noticeable decline in performance} no matter adapting the model to a dissimilar task (e.g., Sync.3m and Skiing are quite not similar) or even a similar one (e.g., both Sync.3m and Single.10m are diving-like actions).

\begin{figure*}[t]
  \centering
   \includegraphics[width=0.85\linewidth]{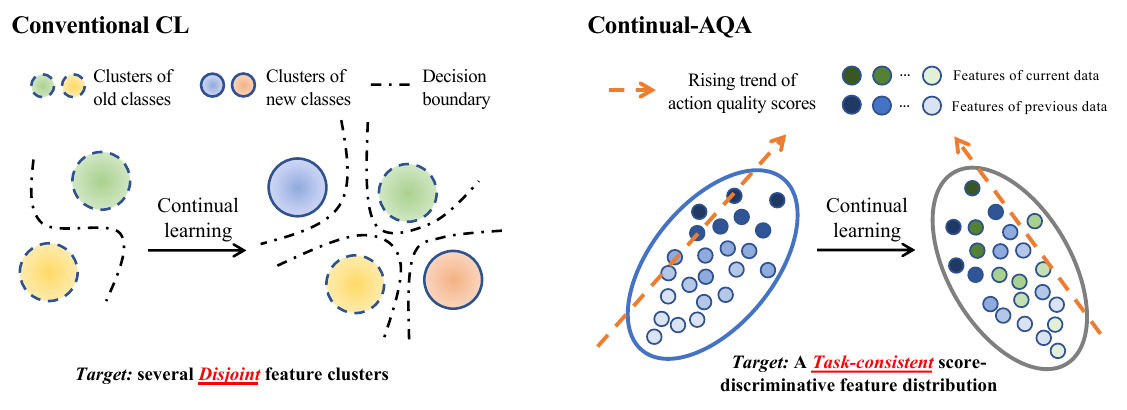}
   \caption{\textbf{Different learning targets of Conventional Continual Learning and Continual-AQA.}
   The learning target of conventional Continual Learning (left) is to construct new decision boundaries so that intra-class features can be clustered and inter-class features can be separated. Differently, the learning target of Continual-AQA (right) is to learn a \emph{task-consistent score-discriminative feature distribution} where the features strongly correlate with the score labels regardless of the task or action types.
   }
   \label{fig:intro_ourWork}
\end{figure*}

To mitigate the forgetting, we \ym{begin with an analysis of the difference} learning targets between Conventional CL and Continual-AQA. 
As shown in Figure~\ref{fig:intro_ourWork}, we claim that to solve Continual-AQA, the key is to sequentially construct a \emph{\textbf{task-consistent score-discriminative feature distribution}} across all seen tasks rather than learning disjoint feature clusters and decision boundaries for \ym{distinct} classes or tasks in conventional CL \cite{icarl, ACL, prototype-aug}.
A further explanation of this claim is that if an AQA model can map \ym{action videos} of seen tasks to the feature space, \ym{where} the latent features express a strong correlation with the score labels regardless of the task or action types, it would be able to handle various tasks without forgetting.
Moreover, considering different AQA tasks have general and specific assessing criteria (e.g., body balance and motion fluency are general criteria; landing distance and splash are specific criteria for Snowboarding and Diving, respectively), it is challenging to extract task-consistent score-discriminative features without explicitly modeling such general and specific knowledge.
Therefore, to achieve Continual-AQA with less forgetting, 
we raise the two following problems: 
1) \emph{How to \textbf{fuse rather than isolate} the feature distribution \ym{learned from current task} with that learned \ym{from the previous tasks}}?
2) \emph{How to explicitly model \textbf{general and specific knowledge} across tasks so that consistent score-discriminative features can be better extracted}? 

% 介绍方法
To address the \ym{aforementioned} problems, we propose a novel \textbf{Feature-Score Correlation-Aware Rehearsal (FSCAR)} to \ym{efficiently utilize} data from previous tasks \ym{within the constraint of} limited memory size. 
First, to sample representative exemplars, an intuitive but effective Grouping Sampling strategy is adopted to sample exemplars that cover different score levels. 
\ym{Additionally, to address the imbalance between previous and current data \cite{prototype-aug,large-scale-IL, ucir}, we propose Feature-Score co-Augmentation to augment the latent features and the scores of previous data. }
Specifically, we introduce the concept of ``augmentation helpers", which provide information on the feature distribution of previous data so that the augmentation can be guided and the augmented feature distribution can correctly reflect the correlation between features and action quality scores.
Moreover, an alignment loss based on difference modeling \cite{gart} is adopted to align the features across tasks and construct a task-consistent score-discriminative feature distribution.

Furthermore, we separately model the action-general and action-specific knowledge to \ym{enhance the} \ym{extraction of task-}consistent score-discriminative features across different tasks). 
Specifically, we build an \textbf{Action General-Specific Graph (AGSG)} over the original joint relation graph proposed in \cite{JRG}. AGSG decouples the original joint relation graph into two parts (i.e., Action General Graph and Action Specific Graph) to model action-general and action-specific knowledge, respectively. We observe that such a design can further mitigate forgetting. 
% Compared with existing AQA works considered the transfer knowledge across AQA tasks, 

In summary, the contributions of our work are as follows:
1) To our best knowledge, this is the first work to study Continual-AQA.
2) We propose a novel Feature-Score Correlation-Aware Rehearsal (FSCAR) to store and reuse data from previous tasks. 
FSCAR distinguishes itself from existing rehearsal methods by its awareness of feature-score correlations.
3) We propose to learn and decouple action-general and action-specific knowledge with an Action General-Specific Graph module for learning different AQA tasks sequentially with less forgetting.

Extensive experiments are conducted, and the results show that: 
1) Compared with directly fine-tuning the model sequentially, our approach explicitly mitigates the forgetting in Continual-AQA. 
2) Our approach outperforms other possible variants, the existing general CL approaches \cite{ewc,icarl,afc,podnet} and successful AQA methods \cite{usdl, gart} trained with naive CL solutions. 
3) Our approach has versatility in various Continual-AQA scenarios.

\section{Related Works}
\textbf{Action Quality Assessment (AQA)} aims to evaluate and quantify the overall performance or proficiency of human actions based on video or motion data analysis.
Most works in AQA mainly focus on problem formulations \cite{whos-better, usdl, gart, xu2022likert,mtl-aqa}, designing effective feature extractors \cite{zeng2020hybrid, best, JRG, tpt,tsa-net,hgcn_aqa_csvt23, learning2score_csvt19,jain2020action}, data insufficiency \cite{semi_supervised_aqa,fitness_aqa}, and so on.
However, existing methods only consider training the model at one time, ignoring the problem of training a model continually. 
In this paper, considering the characteristics of Continual-AQA, we attempt to find a feasible solution for this problem. 

The closest to our work is \cite{JRG} because our AGSG is built on the proposed JRG module. However, the original JRG is not designed for Continual Learning, and AGSG differs from it in graph \ym{construction} and model training. Additionally, a concurrent work \cite{transfer_aqa} considers the transferable knowledge across various tasks and tries to model such knowledge explicitly, but it assumes that data from different tasks are all available during training, which contradicts the invisibility of previous data in Continual-AQA.

\vspace{+0.3cm}
\textbf{Continual Learning (CL)} is a rising research direction to study how to train deep neural networks sequentially. The main problem in CL is that the model will forget the previously learned knowledge when learning new knowledge, namely catastrophic forgetting.

To address catastrophic forgetting, {Rehearsal-based} methods propose to save and reuse several samples of older tasks. 
In the past few years, many methods \cite{icarl, prototype-aug,large-scale-IL, ucir, podnet} focused on selecting representative exemplars and balancing new and limited old data.
However, compared with convention classification-based CL, Continual-AQA has a totally different learning target, which makes the existing methods \cite{icarl,prototype-aug,large-scale-IL, ucir} unsuitable for Continual-AQA with their classification-based designs. 
In this work, a novel rehearsal-based method is proposed considering the distinct learning target explained in Figure~\ref{fig:intro_ourWork}.
\ym{The most related works are iCaRL \cite{icarl} and PASS \cite{prototype-aug}. 
The former one proposes to store limited instances of previous tasks, but the class-clustered based sampling strategy cannot sample the representative exemplars in Continual-AQA.
The latter one proposes to randomly augment the stored features with a Gaussian noises, ignoring the subtle relation between features and scores inter- and intra-AQA tasks.
The Feature-Score Correlation-Aware Rehearsal overcome the weakness of these methods by proposing a grouping sampling to sample representative exemplars and a feature-score co-augmentation to model the subtle relation across stored instances and thier corresponding scores so that the augmentation can be guided.}

Recently, inspired by Complementary Learning \cite{Complementary_learning1,Complementary_learning2}, some works \cite{dualprompt, ACL} propose to model task-invariant and task-specific knowledge to address catastrophic forgetting. 
However, existing works either heavily rely on pre-training \cite{dualprompt} or attempts to construct a disjoint representation for the private knowledge of each task \cite{ACL}. 
In this work, we address Continual-AQA without any pre-training on other AQA tasks. Additionally, we design an Action General-Specific Graph module to extract general and specific knowledge to map the input video actions from different tasks to the features in a task-consistent action quality latent space.

\section{Problem Formulation}
We first review the general setting of AQA: given an action video contains \ym{a single} action (e.g., diving or skiing), the \ym{objective for} the model is \ym{to accurately} predict the quality score of the action. Following existing works \cite{JRG_ASS,JRG,zeng2020hybrid,learning2score_csvt19}, we formulate AQA as a regression-based task.

Now considering a sequence of AQA tasks $D=\{D^1,..., D^T\}$, each task $D^t$ \ym{comprises} $N^t$ action videos $v^t$ with their corresponding action quality scores.
\ym{These tasks, encompassing a variety of action types (e.g., Diving and Snowboarding), arrive sequentially.
Upon the arrival of a new task, the model undergoes training on it.}
It is noteworthy that when training the model on a new task $D^t$, data from previous tasks will not be available, or just a tiny amount (less than a predefined small number $M$) of them can be stored and reused.

The goal of Continual-AQA is to train a unified model $F_\theta: V \to S $ parameterized by $\theta$, so that it can predict the action quality score $\hat{s}=F_\theta(v) \in S $ correctly \ym{for} an unseen test action \ym{video} $v$ from any task seen before. In this work, we separate the model into a feature extractor $E$ and a lightweight score regressor $R_s$, which means $\hat{s}=F_\theta(v)=R_s(E(v))$.

\begin{figure*}[t]
  \centering
%   \fbox{\rule{0pt}{2in} \rule{0.9\linewidth}{0pt}}
   \includegraphics[width=0.9\linewidth]{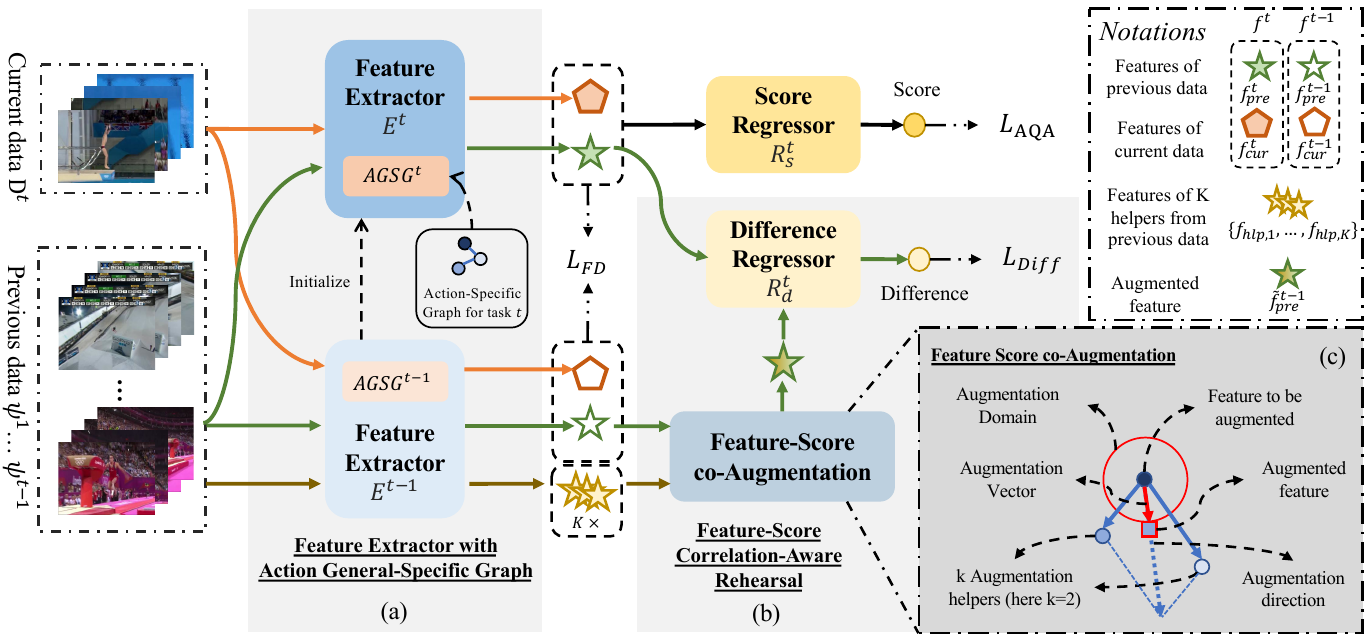}
    % \vspace{-0.3cm}
   \caption{\textbf{A pipeline of our method for Continual-AQA.} Our framework contains two main components: Feature-Score Correlation-Aware Rehearsal(FSCAR) in sub-image (b) and Feature Extractor with Action General-Specific Graph(AGSG) in sub-image (a). Notably, when a new task comes, the model is initialized by the previously learned weights, and a new Action-Specific Graph will be defined for the new task before training. An intuitive example of Feature-Score co-Augmentation is shown in sub-image (c), where a feature with a darker color has a higher quality score. Arrows in different colors indicate different data streams. $\{1,...,t\}$ is the index of tasks and here we \ym{regard} task $t$ \ym{as} the current task and task $t\text{-}1$ \ym{as} the nearest previous task. $D^t$ denotes all the training data in current task, and $\psi=\{\psi^1,...,\psi^{t-1}\}$ denotes the stored exemplars of previous tasks.
   Best viewed in color.}
   \label{fig:framework}
   % \vspace{-0.3cm}
\end{figure*}

\section{Methodology}
\subsection{Overview of Our Method}  
\label{sec:overview}
As illustrated in Section \ref{sec:intro}, Continual-AQA has a different learning target from conventional Classification-based Continual Learning, that is, learning a task-consistent score-discriminative feature distribution across all seen tasks. Motivated by this, two main questions have been raised: 
1) How to fuse rather than isolate the feature distribution \ym{learned from previous tasks} with that learned \ym{from the new task}?
2) How to \ym{explicitly} model general and specific knowledge \ym{across tasks} so that \ym{task-consistent} score-discriminative features can be better extracted?
To address the two questions, we develop our method from the following two aspects.

For the first question, we propose a novel rehearsal-based method, namely \textbf{Feature-Score Correlation-Aware Rehearsal (FSCAR)}, which includes: 1) a Grouping Sampling strategy to sample representative exemplars; 2) a Feature-Score co-Augmentation to augment the previous data so that the imbalance problem can be addressed; 3) a different modeling-based feature alignment loss function $\mathcal{L}_{Diff}$ to align the feature of previous and current data into a task-consistent feature distribution.

\begin{figure}[t]
  \centering
   \includegraphics[width=0.85\linewidth]{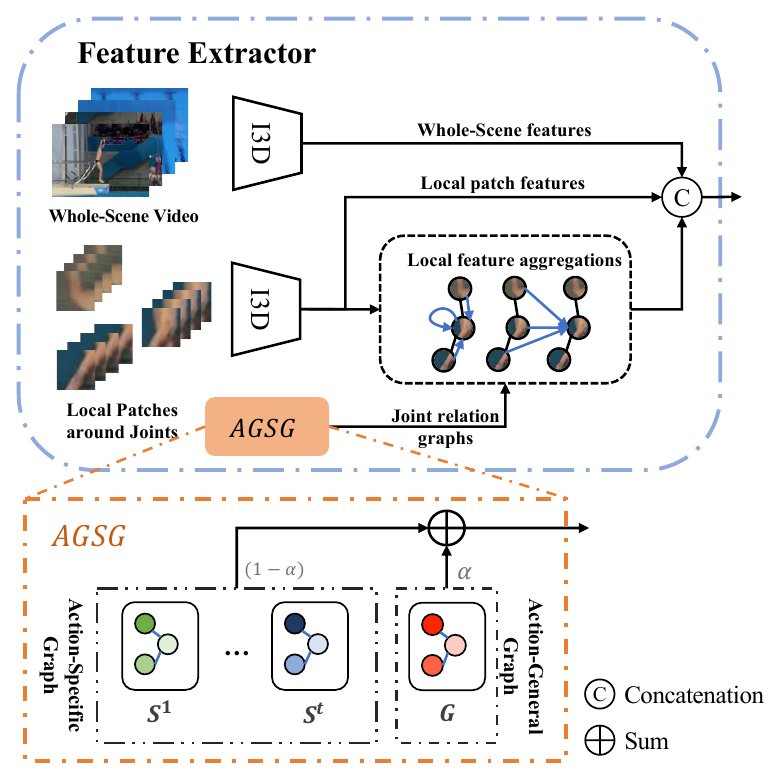}
   \caption{\textbf{Feature extractor with the proposed Action General-Specific Graph (AGSG)}. Different from the original JRG \cite{JRG}, in this work, we decouple the original joint relation graph into an Action-General Graph and an Action-Specific Graph to learn action-general and action-specific knowledge, respectively (the lower part). 
   }
   \label{fig:AGSG}
\end{figure}

For the second question, we propose \textbf{Action General-Specific Graph (AGSG)}, which can be shown as the lower part of Figure~\ref{fig:AGSG}, to strengthen the ability to extract task-consistent features of the feature extractor during continual training. 
AGSG contains an Action-General Graph and an Action-Specific Graph to learn action-general and action-specific knowledge, 
respectively.
Moreover, a distillation-based loss function $\mathcal{L}_{FD}$ is proposed for AGSG to maintain and decouple the already learned knowledge during continual training.

Figure~\ref{fig:framework} illustrates the overall framework of our method.
In the following part of this section, we will introduce FSCAR and AGSG in Section \ref{sec:fs-aug} and Section \ref{sec:AGSG}, respectively. After that, the details about continual training strategies will be introduced in Section~\ref{sec:training}.

\subsection{Feature-Score Correlation-Aware Rehearsal}
\label{sec:fs-aug}
\ym{Our goal is to ensure that the model can accurately represent the feature distribution of previous tasks using a limited stored data. This capability is crucial for integrating new data into this established distribution, resulting in a task-consistent, score-discriminative feature distribution. Such a distribution enables the regressor to reliably predict scores regardless of action types.}
To achieve this, we propose a Feature-Score Correlation-Aware Rehearsal, which \ym{is designed to tackle} two problems, \ym{i.e.}, ``what to replay" and ``how to replay" for Continual-AQA.

\ym{Upon completing} training on task ${t\text{-}1}$, we adopt a \textbf{Grouping Sampling (GS)} strategy to sample representative exemplars of task $t\text{-}1$. 
Suppose we need to sample $m$ exemplars to store in the memory of size M and reuse them in the future training stages. 
GS first uniformly separates data into $m$ groups based on the score ranking. 
\ym{Subsequently,} one exemplar will be sampled from each group. 
The sampled exemplars of task $t\text{-}1$ are \ym{noted} as $\psi_{t-1}$.
Although such a sampling strategy is intuitive, experiments in Section~\ref{sec:ablation_FSDAR} show that it can already roughly describe the relationship between video action instances and quality scores and cover exemplars with different-level scores for previous data. 

To reuse the stored data and solve the imbalance problem \cite{prototype-aug,large-scale-IL,ucir}, when training on task $t$ ($t>1$), the stored exemplars of previous tasks $\{\psi^1,...,\psi^{t-1}\}$ are fed into feature extractor $E^t$ and $E^{t-1}$. 
Suppose $f_{pre,j}^t$ and $f_{pre,j}^{t-1}$ denote the features of a \emph{previous instance} extracted by $E^t$ and $E^{t-1}$\ym{, where $j$ indicates the index of features of stored previous instances}.
Then, a \textbf{Feature-Score co-Augmentation(FS-Aug)} is applied on $f_{pre,j}^{t-1}$ and its corresponding action quality score $s_{pre,j}$ to obtain the augmented feature $\tilde{f}^{t-1}_{pre,j}$ and score $\tilde{s}_{pre,j}$. 
\ym{Specifically}, given a feature to be augmented and its corresponding action quality score, FS-Aug first randomly selects K feature-score pairs from \emph{previous data} as the augmentation helpers, which can be \ym{noted} as $H_{feat}=\{f_{hlp,1}^{t-1},..., f_{hlp,K}^{t-1}\}$ and $H_{score}=\{s_{hlp,1},..., s_{hlp,K}\}$. 
Then, the feature perturbation $\Delta f$ and the score perturbation $\Delta s$ are computed as:

\begin{equation}
  \Delta f_j = \frac{1}{K}\sum_{k=1}^K f^{t-1}_{hlp,k}-f_{pre,j}^{t-1} ,
  \label{eq:delta_f}
\end{equation}
\begin{equation}
  \Delta s_j = \frac{1}{K}\sum_{k=1}^K s_{hlp,k}-s_{pre,j} .
  \label{eq:delta_s}
\end{equation}
Given the perturbations, the augmented feature and score can be computed as:
\begin{equation}
  \tilde{f}^{t-1}_{pre,j}=f_{pre,j}^{t-1} + \varepsilon\Delta f_j ,
  \label{eq:f_after_aug}
\end{equation}
\begin{equation}
  \tilde{s}_{pre,j}=s_{pre,j} + \varepsilon\Delta s_j ,
  \label{eq:s_after_aug}
\end{equation}
where $\varepsilon$ is sampled from a Gaussian distribution $N(0, \sigma)$ and $\sigma$ is a hyper-parameter.

\ym{After augmenting the features and scores of previous data, a remaining question is: \emph{how to utilize the augmented features and scores?} 
To address this issue, our insight is, if the model can learn without forgetting, it should be easy to predict the score different between $\tilde{f}^{t-1}_{pre,j}$ and $ f_{pre,j}^{t}$. Hence, an auxiliary loss is adopted to align the features of previous and current data into a consistent distribution. }
Specifically, inspired by the idea of score difference modeling proposed in CoRe \cite{gart}, an auxiliary difference regressor $R^t_d$ is utilized to predict the score difference between $\tilde{f}^{t-1}_{pre}$ and $f_{pre}^{t}$, and the difference loss is computed as:
\begin{equation}
  d=\tilde{s}_{pre} - s_{pre} ,
  \label{eq:diff}
\end{equation}
\begin{equation}
  L_{Diff}=(R^t_d(\tilde{f}^{t-1}_{pre,j}\copyright f_{pre,j}^{t})-d)^2 ,
  \label{eq:L_diff}
\end{equation}
where $\copyright$ denotes the concatenation operation.

\vspace{0.13cm}
\noindent \textbf{- Discussions.} It has been emphasized in previous works \cite{prototype-aug,large-scale-IL,ucir} that the imbalance of previous and current data will cause forgetting. Hence, inspired by PASS \cite{prototype-aug}, besides training $R^t_s$ with current data and stored previous data, we propose FS-Aug to augment previous data to address the imbalance issue.

We have to claim that augmenting the corresponding scores together with the features is necessary because of the strong correlation between features and scores, which means a subtle perturbation in features would cause a perturbation in scores. 
Noting that PASS \cite{prototype-aug} augments feature by adding random Gaussian noise, it is impossible for PASS to perceive such a strong correlation.
Differently, in our FS-Aug, augmentation helpers are introduced to provide information on the subtle correlation between latent features and scores so that the augmentation can be guided.

\ym{The deeper motivation of Equation 1-4 is also based on the strong correlation between features and scores. Imagine two features $f_i$ and $f_j$ in the latent space (corresponding action quality scores are \ym{noted} as $s_i$ and $s_j$, and $s_i > s_j$), some feature point $f_z$ around the connection between $f_i$ and $f_j$, with high possibility, would corespond to a score $s_z$ which satisfies $s_i > s_z > s_j$. Hence, we can obtain the augmentation direction of $f^{t-1}_{pre,j}$ and score $s^{t-1}_{pre,j}$ with the help of other features (i.e., \emph{helpers}) in the same latent space and its corresponding scores.}

\ym{Based on FS-Aug}, we adopt a difference modeling-based loss to maintain the score-discriminability of $\tilde{f}^{t-1}_{pre,j}$ and $f_{pre,j}^{t}$ given any $i$ so that the forgetting can be constrained.
Ablation studies in Sec.~\ref{sec:ablation_FSDAR} show that our approach outperforms the augmentation method proposed in PASS \cite{prototype-aug} and other possible strategies.

\subsection{Action General-Specific Graph}
\label{sec:AGSG}
\ym{As discussed in Section I, different AQA tasks may have different assessment criteria, which poses a challenge to extracting task-consistent score-discriminative features. Inspired by Complementary Learning Systems \cite{Complementary_learning1, Complementary_learning2}, we believe if we can use a shared action-general module to model shared information across tasks and several specific modules for each task to model action-specific information, task-consistent score-discriminative features can be easier to extract and the forgetting in Continual-AQA will be further mitigated.}
Therefore, we build our general-specific knowledge Modeling module, \ym{termed the} Action General-Specific Graph (AGSG), on JRG \cite{JRG} to learn and decouple action-general and action-specific knowledge so that task-consistent features can be better extracted. 
The pipeline of AGSG can be shown in Figure~\ref{fig:AGSG}.

Following \cite{JRG}, we define a joint relation graph as a learnable weight matrix $A \in \mathbb{R}^{J\times J}$, where $J$ denotes the number of joints. An element $A[i,j]$ denotes the importance of the relationship between the $i$-th and $j$-th joint for assessing one type of action.
In our work, AGSG contains a set of Action-Specific Graphs (ASG) and an Action-General Graph (AGG). An ASG is defined and learned for a single task, and the AGG is shared across different tasks during continual training. Given a task $t$, the AGSG can be computed as:
\begin{equation}
  A^t = \alpha \cdot G + (1-\alpha) \cdot S^t,
  \label{eq:sum_graph}
\end{equation}
where $\alpha \in [0,1]$ is a hyperparameter, $G$ denotes the AGG and $S^t$ denotes the ASG. 
\ym{We will provide more details about feature extractor with AGSG in the Appendix.}

As to the optimization of AGSG, there are two aspects to be considered. 
First, it is expected that the newly learned feature extractor can output similar features as the previously learned one to maintain the learned knowledge.
Second, the action-general and action-specific knowledge are expected to be decoupled during continual training.
To address these, we do not freeze AGG and ASG during the continual training stages and conduct a distillation-based loss function $\mathcal{L}_{FD}$ on the output features obtained from $E^t$ and $E^{t-1}$. 
Specifically, given a video instance of $v^\epsilon_i$ from a seen but agnostic task $\epsilon$, we use the features extracted by $E^{t-1}$ as the distillation target and $\mathcal{L}_{FD}$ can be written as: 
\begin{equation}
    \mathcal{L}_{FD} = \sum^{t-1}_{g=1}||E^t_g(v^*_i) - E^{t-1}_g(v^*_i)||^2.
  \label{eq:reformulate_fd}
\end{equation}
Here we introduce the subscript $g \in \{1,...,t-1\}$ for $E^t$ and $E^{t-1}$ to \ym{represent} that $A^g$ is utilized to extract features. Although those features extracted when using AGSGs except $A^\epsilon$ are meaningless for the inference stage, inspired by LwF \cite{lwf}, we believe they are valuable as pseudo labels for distillation.

\ym{Notably,} it is essential to keep the model initialized in an identical way for fair comparisons, and a better initialization can help a model learn better. Hence, at the base step ($t\text{=}1$), we initialize the ASG to keep the initial $A^1$ the same as those in the model without ASG. At continual training steps, we initialize $S^t$ by the previously learnt ASG at task $t\text{-}1$.
Experiments in Sec.~\ref{sec:ablation_agsg} show that such initialization can gain performance improvement.

\vspace{0.13cm}
\noindent{\textbf{- Discussions.}}
{We develop a stronger feature extractor rather than directly adopting the original JRG \cite{JRG} for the following two reasons. First, the importance of the relationship between joint pairs may differ among actions, for which using a single graph to model joint relationships across different actions is not feasible for Continual-AQA.
Second, considering the general and specific judging criteria across different tasks, action-general and action-specific knowledge should be explicitly modeled.
Hence, AGSG is proposed to model such dual knowledge during continual training. We believe modeling and decoupling the general and specific knowledge across various tasks would further mitigate forgetting because the general knowledge can be shared, and the specific knowledge can be isolated so that task-consistent features can be better extracted. 
Notably, the AGSG module has two-fold novelties: 1) Compared with the existing works on AQA, AGSG is the first module explicitly models general and specific knowledge. 
2) Compared with the original JRG, AGSG is a CL-based module that differs from JRG in graph \ym{construction} and model training.}

\subsection{Continual Training Strategy}
\label{sec:training}
As introduced in \ref{sec:fs-aug}, FS-Aug directly augments $f_{pre,j}^{t-1}$ and $s_{pre,j}$. Therefore, instead of mixing the previous and current data together, given a mini-batch size $b$ for current data, we additionally sample a mini-batch from previous data to compute $f_{pre,j}^{t}$ and $f_{pre,j}^{t-1}$. Moreover, during training, given an instance $v^\epsilon_i$ from task $\epsilon$, $f^{t}$ and $f^{t-1}$ introduced in Section~\ref{sec:fs-aug} are extracted using the AGSG $A^\epsilon$.

To sum up, the complete loss function during training can be written as:
\begin{equation}
    \mathcal{L} = \mathcal{L}_{AQA} + \lambda_{FD} \mathcal{L}_{FD} + \lambda_{Diff} \mathcal{L}_{Diff} ,
  \label{eq:full_loss}
  % \vspace{-0.2cm}
\end{equation}
where $\lambda_{FD}$ and $\lambda_{Diff}$ are hyper-parameters. Following previous works \cite{JRG,mtl-aqa}, AQA loss is defined as the MSE loss across the predicted and ground-truth scores. Note that when training the model on the first task, the later two losses are not computed to optimize the model. For more training details, please refer to the Appendix.

\section{Experiments}
In this section, we first introduce the datasets and evaluation metrics for Continual-AQA,  then describe our implementation details. After that, we perform ablation studies to analyze our model in detail. Finally, comparisons are conducted between the proposed method and other baselines.

% \vspace{0.13cm}
\subsection{Datasets and Evaluation Metrics}
\label{sec:dataset_metric}
\noindent{\textbf{{- Continual-AQA Datasets.}}} 
We conduct experiments to evaluate our model on three public action quality assessment datasets: the AQA-7 dataset \cite{aqa7}, the MTL-AQA dataset \cite{mtl-aqa}, and the BEST dataset \cite{best}. The introductions of the datasets are as follows:

\vspace{0.1cm}
\begin{itemize}
\ym{\item \emph{AQA-7}} \cite{aqa7}: The AQA-7 dataset contains 1,189 videos from seven summer-  and winder-Olympic Games: single diving-10m platform (Diving), gymnastic vault (Gym vault), synchronous diving-3m springboard (Sync.3m), synchronous diving-10m platform, big air skiing (Skiing), big air snowboarding (Snowboard) and Trampoline. Following the previous works \cite{gart, JRG, usdl}, only the videos from the former six games are used in our work.

\vspace{0.1cm}
\ym{\item \emph{MTL-AQA}} \cite{mtl-aqa}: The MTL-AQA dataset contains 1,412 videos from 16 different world diving events. As for the annotations, the MTL-AQA contains various annotations, including difficulty, action type, scores from each judge, final scores, and video captions. Only the final scores are used in our work.

\vspace{0.1cm}
\ym{\item \emph{BEST}} \cite{best}: The BEST is a long video dataset with annotations about the relative rankings of video pairs. Detailly, the BEST dataset contains 16,782 video pairs from five various daily actions (i.e., applying eyeliner, braiding hair, folding origami, scrambling eggs and tying a tie). 
\end{itemize}

\vspace{0.1cm}
\noindent{\textbf{{- Common AQA Metrics.}}} 
In previous works \cite{JRG_ASS, gart, xu2022likert, best, whos-better}, \emph{Pairwise Accuracy} and \emph{Spearman’s rank correlation coefficient (SRCC)} are utilized to evaluate the performance of the model on an AQA task. 
The Pairwise Accuracy represents the percentage of correctly ranked pairs. 
The SRCC can evaluate the correlation between the score series predicted by the model and the ground-truth score series, which is defined as follows: 
\begin{equation}
\rho = \frac{\sum^N_{i=1} (x_i-\bar{x})(y_i-\bar{y})}{\sqrt{\sum^N_{i=1} (x_i-\bar{x})^{2}\sum^N_{i=1}(y_i-\bar{y})^2}},
\label{rho}
\end{equation}
where $x$ and ${y}$ are the rankings of the two series.

Following previous works \cite{JRG_ASS, mtl-aqa, best}, we use SRCC when the experiments are conducted on the AQA-7 and the MTL-AQA datasets and pairwise accuracy when experiments are conducted on the BEST dataset.

\vspace{0.1cm}
\noindent{\textbf{{- Continual-AQA Metrics.}}}
In this work, we utilize Average Performance ($\mathcal{AP}$), Negative Backward Transfer ($\mathcal{NBT}$), and Maximum Forgetting ($\mathcal{MF}$) to evaluate the Continual-AQA models. 
After training $T$ tasks, an upper triangular performance matrix $P$ can be obtained, and the three metrics can be computed as:
\begin{equation}
    \mathcal{AP} = Average(P_{T,1},...,P_{T,T}),
  \label{eq:ap_rho}
\end{equation}
\begin{equation}
    \mathcal{NBT} = \frac{1}{T-1}\sum^{T-1}_{t=1} max(0, P_{t,t}-P_{T,t}),
  \label{eq:bf_rho}
\end{equation}
\begin{equation}
    \mathcal{MF} = \frac{1}{T-1}\sum^{T-1}_{t=1} \max_{i,j\in \{1,...,T\}}(P_{i,t}-P_{j,t}), \forall j \ne t,
  \label{eq:mf_rho}
\end{equation}
where $P_{i,j}$ denotes the performance on task $j$ after training on task $i$.

$\mathcal{AP}$ evaluates the average performance on all seen tasks, and a higher value indicates better performance. $\mathcal{NBT}$ and $\mathcal{MF}$ evaluate the forgetting during training, and a lower value indicates better performance.

\subsection{Implementation Details}
\label{sec:impl_details}

\noindent{\textbf{- Data Preprocessing.}}
For the AQA-7 dataset, we separate it into six non-overlapping subsets according to the type of actions to conduct experiments in a multi-stage training manner. 
We follow \cite{JRG_ASS} to extract I3D \cite{i3d} features from raw videos and optical flows in advance.

For the MTL-AQA dataset, we separate it into two subsets (i.e., 3m-springboard and 10m-platform) based on the type of action, which means the number of tasks is 2. 
Note that under this setting, in the experiments, the values of $\mathcal{NBT}$ are always equal to the values of $\mathcal{MF}$. Hence, only the results on $\mathcal{AP}$ and $\mathcal{NBT}$ are reported in Table 8. 
We follow \cite{semi_supervised_aqa} to extract I3D \cite{i3d} features from raw videos and optical flows in advance.

For the BEST dataset, we separate it into five non-overlapping subsets according to the type of actions and directly use the features extracted by the original work \cite{best}.

\vspace{0.1cm}
\noindent{\textbf{- Model Details.}}
For experiments on the AQA-7 dataset, We follow \cite{JRG} to build our model except for the graph \ym{construction}.
We separate the original model into two parts: a feature extractor and a score regressor.
The difference regressor in our model has a similar architecture as the score regressor except for the input dimension. 

For experiments on the MTL-AQA dataset,  
We adopt an I3D-LSTM as the feature extractor. The I3D backbone is initialized by the pre-trained weights on the Kinetic-400 dataset \cite{k400} and frozen during training. Two layers of LSTMs are adopted to aggregate temporal features and trained from scratch. 
We adopt this setting for the following two reasons. First, human joint annotations are not provided in the MTL-AQA dataset, and it is impossible to build the JRG-based feature extractor directly. Second, we attempt to study whether our FSCAR can be plugged into other feature extractors and works well.

For experiments on the BEST dataset, We build our model based on the model proposed for the BEST dataset in \cite{JRG_ASS}. In our work, we replace the original NAS-based \cite{darts} architecture with a JRG-like
module to model the local temporal information.

\vspace{0.1cm}
\noindent{\textbf{{- Training Settings.}}}
In our work, Adam \cite{adam} is used to optimize our model.
We set the learning rate as 0.01 for AGG and ASG and 0.001 for the other parameters in our model with a $10^{-5}$ weight decay. 
For our Grouping Sampling, we sample the first exemplar for the former $m\text{-}1$ groups and the last exemplar for the $m$-th group to cover the highest and lowest scores. 
We set the total memory size $M$ to 30 unless specified otherwise. 
\ym{
Note that such a memory size will not incur significant storage overhead because the I3D backbone is frozen during training, we can directly store the I3D features and score labels of the instances (e.g., it roughly requires a storage overhead of 41MB in our experiments on the AQA-7 datasets). Even if we store the raw videos, 30 videos will not bring significant storage overhead.}
Following iCaRL \cite{icarl}, we gradually remove exemplars in memory to store new exemplars.
To mitigate the influence of specific task orders, for experiments on the AQA-7 and the BEST dataset, we use four random seeds\footnote{Random Seeds: 0, 1, 2, 3} to generate four task sequences. For the MTL-AQA dataset, we conduct experiments on two possible task sequences. The reported results are average performance on different task sequences.

\vspace{0.1cm}
\noindent{\textbf{{- Existing CL Baselines.}}}
In Section~\ref{sec:main_comparison}, we compare the proposed method with four existing CL methods, including  EWC \cite{ewc}, iCaRL \cite{icarl}, PODNet \cite{podnet} and AFC \cite{afc}. The processing details of the baseline models are as follows:

\vspace{0.1cm}
\begin{itemize}
\ym{\item \emph{EWC}} \cite{ewc}:
Following the original EWC, we conduct a regularization term to constrain the change of the parameters learned from previous tasks.

\ym{\item \emph{iCaRL}} \cite{icarl}:
In this work, we only adopt the Herding strategy to construct the exemplar set without utilizing the NME classifier as the head of the network because it does not work on AQA tasks.

\ym{\item \emph{PODNet}} \cite{podnet}:
In this work, we only adopt the POD loss in the original PODNet to balance previous and current knowledge without utilizing the Local-similarity Classifier as the head of the network because it does not work on AQA tasks. 

\ym{\item \emph{AFC}} \cite{afc}:
Following the original AFC, we compute an ``importance" term to adjust the distillation term. In the comparison results on the AQA-7 and BEST datasets, the ``importance" term is conducted on the POD loss in the PODNet. In the comparison results on the MTL-AQA dataset, the ``importance" term is conducted on the feature distillation loss due to the limitation of selection of feature extractor. 
\end{itemize}

We also report results of two other baselines: Finetune (i.e., directly fine-tuning model on each task) and Upper Bound (i.e., jointly training model all all tasks at once).

\vspace{0.1cm}
\noindent{\textbf{{- Existing AQA Baselines.}}}
In Section~\ref{sec:main_comparison}, we also compare the proposed method with two more existing AQA methods (i.e., USDL \cite{usdl} and CoRe-GART \cite {gart}). The introduction\ym{s} and processing details of the baseline models are as follows:

\vspace{0.1cm}
\begin{itemize}
  \ym{\item \emph{USDL}} \cite{usdl}:
  USDL tries to address AQA by reformulating it as a Distribution Prediction problem. In this work, we implement two variants: 1) original USDL, which follows the official implementation to train the model. 2) USDL*, whose I3D backbone is frozen, and only the distribution predictor is trained across tasks.
  \ym{\item \emph{CoRe-GART}} \cite{gart}:
  CoRe-GART proposes to address AQA by reformulating it as a Contrastive Regression Problem, that is, regressing the relative scores between query videos and exemplar videos. In this work, we implement CoRe-GART by following the official implementation.
\end{itemize}

\subsection{Quantitative Results on each Component}
% \vspace{0.1cm}

Above all, we conduct ablation studies to evaluate the contribution of each proposed component. 
The proposed approach is built on a naive model which is optimized on the AQA loss $\mathcal{L}_{AQA}$ and a naive version of feature distillation loss, where

\begin{equation}
    \mathcal{L}_{FD(Naive)}=||f^t-f^{t-1}||^2.
\label{eq:naive_fd}
\end{equation}
When AGSG is used, the $\mathcal{L}_{FD}$ is reformulated as is introduced in Equation~\ref{eq:reformulate_fd}.
As shown in Table~\ref{tab:component_ablation_aqa7}, \ref{tab:component_ablation_best}, and \ref{tab:component_ablation_mtl}, when gradually adopting the proposed components, the performance gets stably improved. Compared with the naive model, our approach significantly improves the average performance and mitigates forgetting of the model on all seen tasks. 
In addition, comparing the performance on three datasets, we notice that the model achieves lower $\mathcal{AP}$, higher $\mathcal{NBT}$ and $\mathcal{MF}$ on the AQA-7 dataset. This is because the large domain gap of tasks in the AQA-7 dataset (e.g., Sync.3m and Skiing) causes higher forgetting, and some challenging tasks (e.g., Snowboard) cause lower average performance. While on the MTL-AQA dataset, the domain gap between two diving-like actions is small, naturally making the model suffer from less forgetting. 
As for the BEST dataset, we note that we use different metrics (i.e., pairwise accuracy rather than SRCC) to calculate the $\mathcal{AP}$, $\mathcal{NBT}$ and $\mathcal{MF}$ and there is no extremely challenging task in the dataset, which make the performance looks better.

\begin{table}[t]
	\centering
	\scriptsize
     \caption{Ablation studies of the proposed components on the AQA-7 dataset.}
	\begin{tabular}{cccc|ccc}
		\toprule  % 顶部线
		$\mathcal{L}_{FD}$ & GS & $\mathcal{L}_{Diff}$ & AGSG & $\mathcal{AP}(\uparrow)$ & $\mathcal{NBT}(\downarrow)$ & $\mathcal{MF}(\downarrow)$  \\ 
		\bottomrule  % 底部线
		\checkmark &   &   &  & 0.5501 & 0.2403 & 0.3169\\
		\checkmark & \checkmark &   &   & 0.6265 & 0.1770 & 0.2209\\
		% \cmidrule(r){1-7}
		\checkmark & \checkmark & \checkmark &   & 0.6404 & 0.1571 & 0.1930\\
		\hline
		\checkmark & \checkmark & \checkmark & \checkmark & \textbf{0.6492} & \textbf{0.1085} &  \textbf{0.1399}\\
		\bottomrule  % 底部线
	\end{tabular}
    \label{tab:component_ablation_aqa7}
\end{table}

\begin{table}[t]
	\centering
	\scriptsize
     \caption{Ablation studies of the proposed components on the BEST dataset.}
	\begin{tabular}{cccc|ccc}
		\toprule  % 顶部线
		$\mathcal{L}_{FD}$ & GS & $\mathcal{L}_{Diff}$ & AGSG & $\mathcal{AP}(\uparrow)$ & $\mathcal{NBT}(\downarrow)$ & $\mathcal{MF}(\downarrow)$  \\ 
		\bottomrule  % 底部线
		\checkmark &   &   &  & 0.6466 & 0.2115 & 0.2428\\
		\checkmark & \checkmark &   &   & 0.7668 & 0.0608 & 0.0870\\
		\checkmark & \checkmark & \checkmark &   & 0.7857 & 0.0372 & 0.0456\\
		\hline
		\checkmark & \checkmark & \checkmark & \checkmark & \textbf{0.7883} & \textbf{0.0288} &  \textbf{0.0366}\\
		\bottomrule  % 底部线
	\end{tabular}
    \label{tab:component_ablation_best}
\end{table}

\begin{table}[t]
	\centering
	\scriptsize
     \caption{Ablation studies of the proposed components on the MTL-AQA dataset.}
	\begin{tabular}{ccc|ccc}
		\toprule  % 顶部线
		$\mathcal{L}_{FD}$ & GS & $\mathcal{L}_{Diff}$ & $\mathcal{AP}(\uparrow)$ & $\mathcal{NBT}(\downarrow)$  \\ 
		\bottomrule  % 底部线
		\checkmark &   &  & 0.7420 & 0.0997 \\
		\checkmark & \checkmark &    & 0.7518 & 0.0801\\
          \hline
		\checkmark & \checkmark & \checkmark   & \textbf{0.7895} & \textbf{0.0574} \\
		
		\bottomrule  % 底部线
	\end{tabular}
    \label{tab:component_ablation_mtl}
\end{table}

\subsection{Analysis on Feature-Score Correlation-Aware Rehearsal}
\label{sec:ablation_FSDAR}
As mentioned before, the Feature-Score Correlation-Aware Rehearsal mainly contains three parts: Grouping Sampling, Feature-Score co-Augmentation, and an auxiliary difference loss. To verify the effectiveness of these parts, we conduct experiments from the following aspects.

\begin{table}[t]
	\centering
	% \tablestyle{7pt}{1.1}
 	\caption{Comparisons with different sampling strategies on the AQA-7 dataset. In this experiment, all the methods are built on our naive model (Row 1 in Table~\ref{tab:component_ablation_aqa7}). RS indicates the random sampling strategy and M indicates the total memory size.}
	\begin{tabular}{c|c|ccc}
		\toprule  % 顶部线
		Methods & M & $\mathcal{AP}(\uparrow)$ & $\mathcal{NBT}(\downarrow)$ & $\mathcal{MF}(\downarrow)$ \\ 
		\hline
		% \hline
		{Herding \cite{icarl}} & \multirow{3}{*}{30} & 0.5811 & 0.2215 & 0.3289  \\
		{RS} & \multirow{3}{*}{} & 0.6018 & 0.1810 & 0.2334  \\
        {Our GS} & \multirow{3}{*}{} & \textbf{0.6265} & \textbf{0.1770} & \textbf{0.2209} \\
		\hline
		{Herding \cite{icarl}} & \multirow{3}{*}{60} & 0.5915 & 0.2068 & 0.2601  \\
		{RS} & \multirow{3}{*}{} & 0.6298 & 0.1584 & 0.2073  \\
        {Our GS} & \multirow{3}{*}{} &  \textbf{0.6417} & \textbf{0.1454}  &  \textbf{0.1906} \\
		\bottomrule  % 底部线
	\end{tabular}
	\label{tab:comparison_replay}
\end{table} 

\begin{figure}[t]
  \centering
   \includegraphics[width=0.8\linewidth]{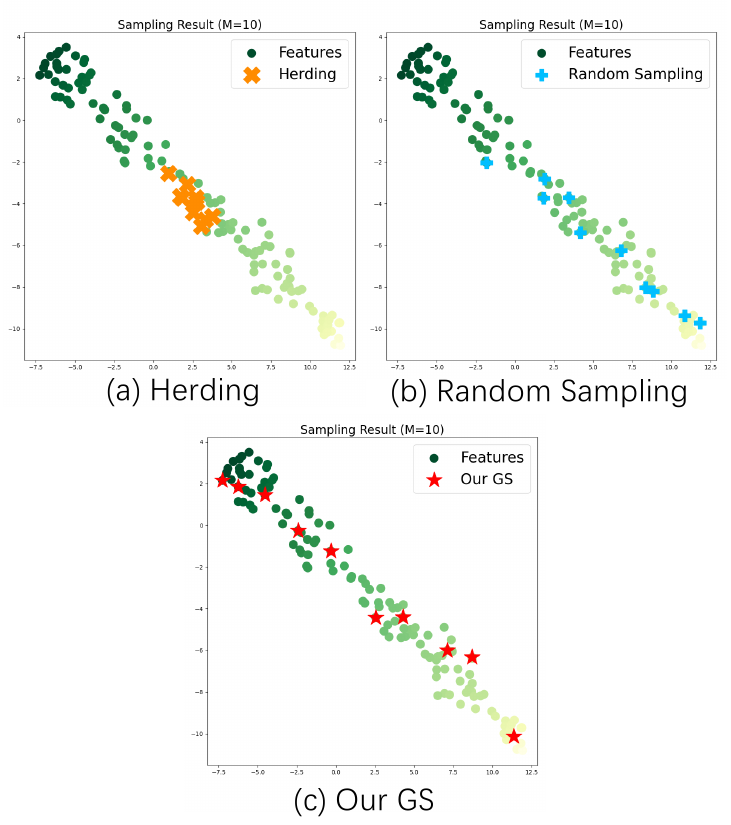}
   \caption{\ym{T-SNE Visualizations of the features sampled by \textbf{Herding (a)}, \textbf{Random Sampling (b)} and \textbf{our GS (c)} after training on the first task. Points in greens denote the features. Darker color indicates higher score. Points in orange, blue and red denote the feature sampled by Herding, Random Sampling and our GS, respectively. \textbf{The exemplars sampled by our GS can cover all score levels and roughly describe the learned feature distribution} rather than gathering in a compact region. Best viewed in color.}  }
   \label{fig:sampled_cpmparison}
\end{figure}

\begin{table*}[htbp]
	\centering
    \caption{Ablation studies about the proposed rehearsal-based paradigm on the AQA-7 dataset. We design three other possible substitutes for comparisons. A variant named ``xx-Aug-yy" indicates we adopt ``xx" (No Augmentation / F-Aug / FS-Aug) to augment features and utilize ``yy" (A-Rgs / A-Diff / A-Dist) to constrain the forgetting. }
	\begin{tabular}{c|cc|ccc|ccc}
		\toprule  % 顶部线
		Variants & F-Aug & FS-Aug & A-Rgs & A-Diff & A-Dist & $\mathcal{AP}(\uparrow)$ & $\mathcal{NBT}(\downarrow)$ & $\mathcal{MF}(\downarrow)$  \\ 
		\hline
		No-Aug-Dist  &   &  &  &  & \checkmark & 0.6261 & 0.1292 & 0.1606 \\
		\hline
		F-Aug-Rgs &\checkmark &   & \checkmark &  &  & 0.6127 & 0.1223 & 0.1532\\
		FS-Aug-Rgs & {} & \checkmark & \checkmark  &  &  & 0.6284 & 0.1197 & 0.1447 \\
		\hline
		Ours & {} & \checkmark &  & \checkmark &  & \textbf{0.6492} & \textbf{0.1085} & \textbf{0.1399}\\
		\bottomrule  % 底部线
	\end{tabular}
	\label{tab:aug-diff_ablation}
\end{table*}

\vspace{0.13cm}
\noindent{\textbf{- Effectiveness of Grouping Sampling.}} In our work, the proposed rehearsal-based method is based on the Grouping Sampling strategy. 
Note that lots of methods \cite{podnet,large-scale-IL,videoIL_iccv21} directly adopt the Herding strategy \cite{icarl} to sample representative exemplars. We conduct a comparison among Herding, 
Random Sampling (RS) and our Grouping Sampling (GS) under two different memory sizes, i.e., 30 and 60.
As shown in Table~\ref{tab:comparison_replay}, 
\ym{when adopting different sampling strategies on our naive model (i.e., Row 1 in Table~\ref{tab:component_ablation_aqa7}),}
Herding gets the worst results, even worse than Random Sampling, and our GS strategy outperforms other methods regardless of the memory size. The results suggest that Herding cannot select the most representative exemplars under Continual-AQA settings. The visualization results shown in Figure~\ref{fig:sampled_cpmparison} also support this view. The features sampled by Herding \ym{Figure~\ref{fig:sampled_cpmparison}(a)} are gathered in a compact region in the latent space. The exemplars' corresponding scores are pretty close, which cannot honestly describe the previously learned feature distribution.
\ym{Additionally, although features with various levels of scores could be sampled by random sampling, still, 
the real data distribution cannot be accurately described, i.e., some exemplars with pretty close scores will be sampled.  Also, data with some levels of scores may be missed by RS, i.e, in Figure~\ref{fig:sampled_cpmparison}(b) no data with high scores is sampled by RS.
On the contrary, our GS Figure~\ref{fig:sampled_cpmparison}(c) can cover all score levels and roughly describe the learned feature distribution rather than gathering in a compact region, which can serve as a general sampling strategy for Continual-AQA.}

\vspace{0.13cm}
\noindent{\textbf{- Impact of Feature-Score co-Augmentation and Difference Loss.}}
The Feature-Score co-Augmentation (FS-Aug) is proposed to augment features and their corresponding scores. Based on FS-Aug, an auxiliary difference loss (A-Diff) is utilized to constrain the shift of already learned feature distribution. Hence, it is worth studying if it is necessary to couple FS-Aug and A-Diff. 
To the end, we consider three other possible methods here:

\begin{itemize}
\item \emph{Feature Augmentation only (F-Aug).} 
As a substitute for FS-Aug, following PASS \cite{prototype-aug}, we augment $f^{t-1}_{pre}$ by adding random Gaussian noises without augmenting their corresponding scores. 
\vspace{0.05cm}
\item \emph{Auxiliary Distillation Loss (A-Dist).}
As a substitute for A-Diff, we directly conduct an extra feature distillation loss on $f^t_{pre}$ and $f^{t-1}_{pre}$.
\vspace{0.05cm}
\item \emph{Auxiliary Score Regression Loss (A-Rgs).}
As a substitute for A-Diff, we perform an extra regression to the scores of the augmented features and adopt an auxiliary regression loss to optimize the model.
\end{itemize}

After that, we combine these methods to build and examine three other variants.
As shown in Table~\ref{tab:aug-diff_ablation}, our approach outperforms other possible strategies on all three metrics. 
The comparison between ``No-Aug-Dist" and our method indicates that an auxiliary distillation loss is ineffective for mitigating forgetting. 
Moreover, compared with ``F-Aug-Rgs", ``FS-Aug-Rgs" performs all the way better, which proves that augmenting corresponding scores together with features is necessary for Continual-AQA.

\vspace{0.13cm}
\ym{\noindent{\textbf{- Impact of the selection of helpers.}}}
We \ym{first} study the impact of the number of helpers.
As shown in Figure 6, the performance improves when the number of helpers grows from 3 to 7. 
  It suggests that when we set an extremely low value to the number of helpers, the perturbation direction obtained by the helpers may not be reliable enough. \ym{In another words, it is necessary to model the feature-score relations between the feature to be augmented and the helpers with various scores.}
  Moreover, the performance achieves slightly decline when further increasing the number of helpers from 7 to 9. 
  \ym{We think this is because when an extremely high value is set to be the number of helpers, the impact of each helper tends to be counteracted due to the opposite direction of the perturbations.} 

\ym{Notably, we use the simplest solution, randomly sampling (noted as \emph{\textbf{Random}}), to select helpers by default.
We beliveve it is desirable to explore if there is a better way to select helpers. To inspire future works, we develop an \emph{\textbf{Anchor-based}} helper selection strategy. We introduce the details of such strategy in Appendix. As shown in Table~\ref{tab:comparison_select_helpers}, the performance gains improvement with the \emph{Anchor-based} helper selection strategy.

Although ``\emph{Anchor-based}'' shows its effectiveness, it requires multiple ranking phases during training, which brings further time cost. Therefore, it is still meaningful to explore better strategies. Also, different helpers may have different importance for augmenting features and scores. We hope future effort will be devoted to these issues to further mitigate the forgetting in Continual-AQA.
}
\begin{table}[t]
	\centering
	% \tablestyle{7pt}{1.1}
 	\caption{Comparison between different helper selection strategies.}
	\begin{tabular}{c|ccc}
		\toprule  % 顶部线
		Variants & $\mathcal{AP}(\uparrow)$ & $\mathcal{NBT}(\downarrow)$ & $\mathcal{MF}(\downarrow)$ \\ 
		\midrule
		% \hline
        {Random} &  {0.6492} & {0.1085} & {0.1399} \\
        {Anchor-based} &  {0.6545} & {0.0903} & {0.1277} \\
		\bottomrule  % 底部线
	\end{tabular}
	\label{tab:comparison_select_helpers}
\end{table} 

\begin{figure}[t]
  \centering
   \includegraphics[width=1\linewidth]{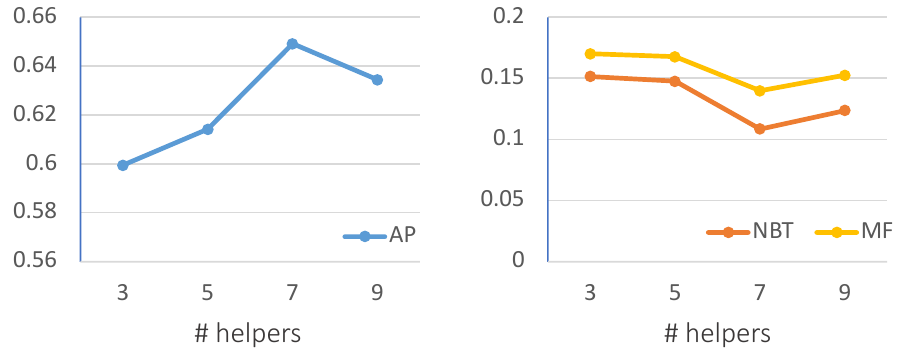}
   \caption{\textbf{The impact of the number of helpers to the Average Performance (AP) and to the Negative Backward Transfer and Maximum Forgetting (NBT and MF) on the AQA-7 dataset.} The X-axis indicates the number of helpers (\# helpers) and {the Y-axis indicates the value of the metrics.}}
   % 图注表示纵坐标
   \label{fig:num_helper_impact}
\end{figure}

\begin{table}[t]
	\centering
 	\caption{Ablation studies of the initialization for Action-specific Graph and the reformulated feature distillation loss on the AQA-7 dataset.}
	\begin{tabular}{c|ccc}
		\toprule  % 顶部线
		Variants & $\mathcal{AP}(\uparrow)$ & $\mathcal{NBT}(\downarrow)$ & $\mathcal{MF}(\downarrow)$ \\
		\hline
		Zero-init & 0.6360 & 0.1565 & 0.1803\\
        Random-init & 0.6401 & 0.1511 & 0.2068\\
        \hline
         w/ $\mathcal{L}_{FD(Naive)}$ & 0.6417 & 0.1360 & 0.1671\\
		\hline
		Ours & \textbf{0.6492} & \textbf{0.1085} & \textbf{0.1399} \\
		\bottomrule  % 底部线
	\end{tabular}
	\label{tab:ablation_AGSG}
\end{table}

\subsection{Analysis on Action General-Specific Graph}
\label{sec:ablation_agsg}
In our work, an Action General-Specific Graph module is developed to learn and decouple the action-general and action-specific knowledge during continual training. As introduced in Section~\ref{sec:AGSG}, we utilize specific initialization for Action-Specific Graph. Moreover, we design a specific feature distillation loss to save and decouple action-general and action-specific knowledge. 
To demonstrate the effectiveness of such designs, on the one hand, we replace the proposed initialization with two intuitive strategies (i.e., zero initialization and random initialization).
On the other hand, we replace the proposed feature distillation loss with the naive feature distillation loss in Equation~\ref{eq:naive_fd}. As shown in Table~\ref{tab:ablation_AGSG}, when ablating any of them, the performance declines significantly, which verifies the contributions of such designs to the performance.

\subsection{Comparison with the Existing Methods on the AQA-7 dataset}
\label{sec:main_comparison}

\begin{table}[t]\small
 \centering
 \footnotesize

 % \tablestyle{10pt}{1.1}
 \caption{Comparisons of our approach with existing AQA methods on the AQA-7 dataset. ``w/ Our GS" denotes our Grouping Sampling is adopted to sample the exemplars. ``w/ PODLoss" denotes the model is trianed together with the POD Loss \cite{podnet}.  USDL* indicates that we frozen the I3D backbone when training USDL model.
 }
    \resizebox{\linewidth}{!}
    {
	\begin{tabular}{l|ccc}
		\toprule  % 顶部线
        % \Xhline{2pt}
		 \multicolumn{1}{l|}{Methods} &   {$\mathcal{AP}(\uparrow)$} & $\mathcal{NBT}(\downarrow)$ & $\mathcal{MF}(\downarrow)$\\
        \hline
        \multicolumn{1}{l|}{\textit{\textbf{AQA model + Finetuning}}} \\ 
        USDL*\cite{usdl} &   {0.2868} & 0.4612 & 0.5573\\
        USDL\cite{usdl} &   {0.3344} & 0.5124 & 0.7373\\
        CoRe-GART\cite{gart} &   {0.3015} & 0.5570 & 0.6336 \\
        % & I3D-MLP & 0.3644 & 0.4341 & 0.5454 \\
        JRG\cite{JRG} &   {0.5523} & 0.2515 & 0.3420 \\
        \hline
        \multicolumn{1}{l|}{\textit{\textbf{AQA model + Continual Learning }}} \\
        % USDL \cite{usdl} w/ PODNet\cite{podnet}\\
        USDL*\cite{usdl} w/ Our GS &   {0.4463} & 0.2844 & 0.3621\\
        USDL\cite{usdl} w/ Our GS &   {0.5852} & 0.2364 & 0.3891\\
        CoRe-GART\cite{gart} w/ Our GS &   {0.4716} & 0.3508 & 0.5391\\
        CoRe-GART\cite{gart} w/ PODLoss\cite{podnet} w/ Our GS  &   {0.4956} & 0.3410 & 0.5108\\
        % \hline
		\cellcolor{yellow}Ours &   \cellcolor{yellow}{\textbf{0.6492}} & \cellcolor{yellow}\textbf{0.1085} & \cellcolor{yellow}\textbf{0.1399} \\
		\bottomrule  % 底部线
      % \Xhline{2pt}
        % \hline
    \end{tabular}
 }
\label{tab:comparison_aqa}
\end{table}

\begin{table}[h]\small
	\centering
 \tablestyle{10pt}{1.1}
 \caption{Comparisons of our approach with existing CL methods on the AQA-7 dataset. ``w/ Our GS" denotes our Grouping Sampling is adopted to sample the exemplars. ``\# Param(m)" indicates the number of trainable parameters.}
    \resizebox{1\linewidth}{!}{
	\begin{tabular}{l|c|ccc}
		\toprule  % 顶部线
		{Methods} & \# Param(m) &$\mathcal{AP}(\uparrow)$ & $\mathcal{NBT}(\downarrow)$ & $\mathcal{MF}(\downarrow)$ \\ 
		\hline
        Finetune & 1.265 &  0.5523 & 0.2515 & 0.3420\\
        Upper Bound & 1.265 & 0.7344 & - & - \\
        \hline
        \textbf{\emph{Train without Memory}} &&\\
        % %\hdashline
		EWC \cite{ewc}& 1.265 & 0.5616 & 0.2734 & 0.4018 \\
        \hline
        \textbf{\emph{Train with Herding}} &&\\
		iCaRL \cite{icarl} & 1.265 & 0.5811 & 0.2215 & 0.3289\\
		PODNet \cite{podnet} &1.265 & 0.5914 & 0.1989 & 0.2530\\
        AFC \cite{afc}  & 1.265 & {0.5787} & {0.1925} & {0.2647}\\ 
        \hline
        \textbf{\emph{Train with Our GS}} & &\\
        % %\hdashline
        PODNet \cite{podnet} w/ Our GS & 1.265 & 0.6393 & 0.1371 & 0.1987\\
        AFC \cite{afc} w/ Our GS & 1.265 & 0.6489 & 0.1275 & 0.1806\\ 
        % \hline
		\cellcolor{yellow}Ours & \cellcolor{yellow} 1.279 (+1.17\%) & \cellcolor{yellow}\textbf{0.6492} & \cellcolor{yellow}\textbf{0.1085} & \cellcolor{yellow}\textbf{0.1399}\\
		\bottomrule  % 底部线
	\end{tabular}
 }
		\label{tab:main_comparison}
\end{table}

\begin{table}[t]
	\centering
	\tablestyle{10pt}{1.1}
    \caption{Comparison results with various total memory sizes (M) on the AQA-7 dataset. ``w/ Our GS" indicates our Grouping Sampling is adopted to sample the exemplars.}
    \resizebox{1\linewidth}{!}{
    	
	\begin{tabular}{l|c|ccc}
		\toprule  % 顶部线
		Methods & M & $\mathcal{AP}(\uparrow)$ & $\mathcal{NBT}(\downarrow)$ & $\mathcal{MF}(\downarrow)$ \\ 
        \hline
        Finetune & - & 0.5523 & 0.2515 & 0.3420 \\
        Upper Bound & - & 0.7344 & - & - \\
        \hline
        {iCaRL} \cite{icarl} & {15} & 0.5602 & 0.2323 & 0.3326  \\
        {PODNet} \cite{podnet} & {15} & 0.5789 & 0.1796 & 0.2699 \\
		{PODNet} \cite{podnet} w/ Our GS & {15} & 0.6013 & 0.1743 & 0.2223  \\
        \cellcolor{yellow}{Ours} & \cellcolor{yellow}{15} & \cellcolor{yellow}\textbf{0.6167} & \cellcolor{yellow}\textbf{0.1272} & \cellcolor{yellow}\textbf{0.1662} \\
        \hline
        {iCaRL} \cite{icarl} & {30} & 0.5811 & 0.2215 & 0.3289  \\
        {PODNet} \cite{podnet} & {30} & 0.5914 & 0.1989 & 0.2530  \\
		{PODNet} \cite{podnet} w/ Our GS & {30} & 0.6393 & 0.1371 & 0.1987  \\
        \cellcolor{yellow}{Ours} & \cellcolor{yellow}{30} & \cellcolor{yellow}\textbf{0.6492} & \cellcolor{yellow}\textbf{0.1085} & \cellcolor{yellow}\textbf{0.1399} \\
		\hline
		{iCaRL} \cite{icarl} & {60} & 0.5915 & 0.2068 & 0.2601  \\
        {PODNet} \cite{podnet} & 60 & 0.6012 & 0.1603 & 0.2259  \\
		{PODNet} \cite{podnet} w/ Our GS & {60} & 0.6621 & 0.1161 & 0.1635  \\
        \cellcolor{yellow}{Ours} & \cellcolor{yellow}{60} &  \cellcolor{yellow}\textbf{0.6712} & \cellcolor{yellow}\textbf{0.0944}&  \cellcolor{yellow}\textbf{0.1367} \\

		\bottomrule  % 底部线
	\end{tabular}
    }
	\label{tab:comparison_memory_size}
\end{table}

We compare the proposed method with the existing methods on the AQA-7 dataset in the following three aspects:

\vspace{0.1cm}
\noindent{\textbf{- Comparison with existing AQA methods.}}
We build C\ym{L} baselines based on successful AQA models (i.e., USDL \cite{usdl}, CoRe-GART \cite{gart}) and compare them with our approach. The comparison results are shown in Table~\ref{tab:comparison_aqa}, and we have the following observations. 
First, when directly fine-tuning models across tasks, each method performs poorly with high forgetting. 
Second, we train USDL and CoRe-GART to build stronger baselines with the POD loss \cite{podnet} and our GS. However, although POD loss and GS significantly improve the performance of USDL and CoRe-GART, our approach outperforms these two methods. 

It is worthwhile noting that the enormous contrast in performance of USDL and CoRe-GART on conventional AQA tasks \cite{usdl, gart} and Continual-AQA may lie in the great task-specific knowledge learning ability with the special problem formulations (e.g., Uncertain Distribution Prediction and Contrastive Regression). 
The more effective to learn specific task, the \ym{easier} to forget the already learned knowledge.

\vspace{0.1cm}
\noindent{\textbf{- Comparison with existing CL methods.}}
We compare the proposed method with four existing CL methods (i.e, EWC \cite{ewc}, icarl \cite{icarl}, podnet \cite{podnet}, and AFC \cite{afc}). 
As shown in Table~\ref{tab:main_comparison}, our method bridges the gap between ``Finetune" and ``Upper Bound" on $\mathcal{AP}$ and significantly alleviates forgetting. 
Additionally, our method outperforms other methods among the three metrics.  
It has been discussed in Section~\ref{sec:ablation_FSDAR} that Herding may not be an appropriate sampling strategy for Continual-AQA. Hence, we further adopt our Grouping Sampling on PODNet and AFC for better comparison.
As shown in Table~\ref{tab:main_comparison}, when our Grouping Sampling is adopted, existing CL methods can perform better but still underperform our approach, further showing our approach's effectiveness.
\ym{Moreover, we calculated the parameters needed for training of different methods and revised TABLE VIII in our manuscript. As shown in Table~\ref*{tab:main_comparison}, we can conclude that \textbf{our method achieves significant improvement with only 1.17\% increase in the number of parameters (from 1.265m to 1.279m)}. Note that in experiments on AQA-7, the parameter increase mainly comes from two aspects. First, the Action-Specific Graphs bring nearly 13.9k additional parameters (less than 2k parameters for each task), and the difference regressor only brings nearly 1.03k additional parameters.}

\begin{table}[t]\small
	\centering
 	\caption{Comparisons of our approach with existing Continual Learning methods on the BEST dataset. ``w/ Our GS" indicates our Grouping Sampling is adopted to sample the exemplars.}
   \resizebox{0.85\linewidth}{!}{
	\begin{tabular}{l|ccc}
		\toprule  % 顶部线
		{Methods} & $\mathcal{AP}(\uparrow)$ & $\mathcal{NBT}(\downarrow)$ & $\mathcal{MF}(\downarrow)$ \\ 
        \hline
        Finetune & 0.6614 & 0.1976 & 0.2381 \\
        Upper Bound & 0.8163 & - & - \\
        \hline
        \textbf{\emph{Train without memory}} \\
        %\hdashline
		EWC \cite{ewc} & 0.6752 & 0.1856 & 0.2137 \\
        \hline
        \textbf{\emph{Train with Herding}} \\
        %\hdashline
		iCaRL \cite{icarl} & 0.7255 & 0.1161 & 0.1378\\
        PODNet \cite{podnet} &  0.7449 & 0.0649 & 0.0699 \\
        AFC \cite{afc} & 0.7566 & 0.0535 & 0.0765 \\
        \hline
        \textbf{\emph{Train with Our GS}} \\
        %\hdashline
        PODNet \cite{podnet} w/ Our GS & 0.7600 & 0.0495 & 0.0708\\
        AFC \cite{afc} w/ Our GS & 0.7643 & 0.0426 & 0.0542 \\ 
		\cellcolor{yellow}Ours & \cellcolor{yellow}\textbf{0.7883} & \cellcolor{yellow}\textbf{0.0288} & \cellcolor{yellow}\textbf{0.0366}\\
		\bottomrule  % 底部线
	\end{tabular}
   }
	\label{tab:comparison_best}
\end{table}

\vspace{0.1cm}
\noindent{\textbf{- Robustness study on memory size.}}
We further evaluate the robustness of our approach by adjusting the total memory size (M). As shown in Table~\ref{tab:comparison_memory_size}, our approach outperforms PODNet \cite{podnet} regardless of the memory size, even if our GS is adopted. 

Moreover, although increasing the memory size can stably improve the performance, there is still an explicit gap between our approach and the joint training upper bound, even if the memory size is set as 60. Such results indicate that there is still room for further research on Continual-AQA.

\begin{table}[t]\small
	\centering
 	\caption{Comparisons of our approach with existing Continual Learning methods on the MTL-AQA dataset. ``w/ Our GS" indicates our Grouping Sampling is adopted to sample the exemplars.}
   \vspace{-0.13cm}
   \resizebox{0.7\linewidth}{!}{
	\begin{tabular}{l|ccc}
		\toprule  % 顶部线
		{Methods} & $\mathcal{AP}(\uparrow)$ & $\mathcal{NBT}(\downarrow)$  \\ 
        \hline
        Finetune & 0.7045 & 0.1697 \\
        Upper Bound & 0.8260 & - \\
        \hline
        \textbf{\emph{Train without memory}} \\
        %\hdashline
		EWC \cite{ewc} & 0.7198 & 0.1607 \\
        \hline
        \textbf{\emph{Train with Herding}} \\
        %\hdashline
		iCaRL \cite{icarl} & 0.7465 & 0.1165 \\
        AFC \cite{afc} & {0.7487} & {0.0979} \\
        \hline
        \textbf{\emph{Train with Our GS}} \\
        %\hdashline
        AFC \cite{afc} w/ Our GS & 0.7690 & 0.0624 \\ 
        % \hline
        \cellcolor{yellow}Ours & \cellcolor{yellow}\textbf{0.7895} & \cellcolor{yellow}\textbf{0.0574}\\
		\bottomrule  % 底部线
	\end{tabular}
   }
	\label{tab:comparison_mtl_aqa}
  \vspace{-0.2cm}
\end{table}

\begin{figure*}[t]
  \centering
   \includegraphics[width=0.9\linewidth]{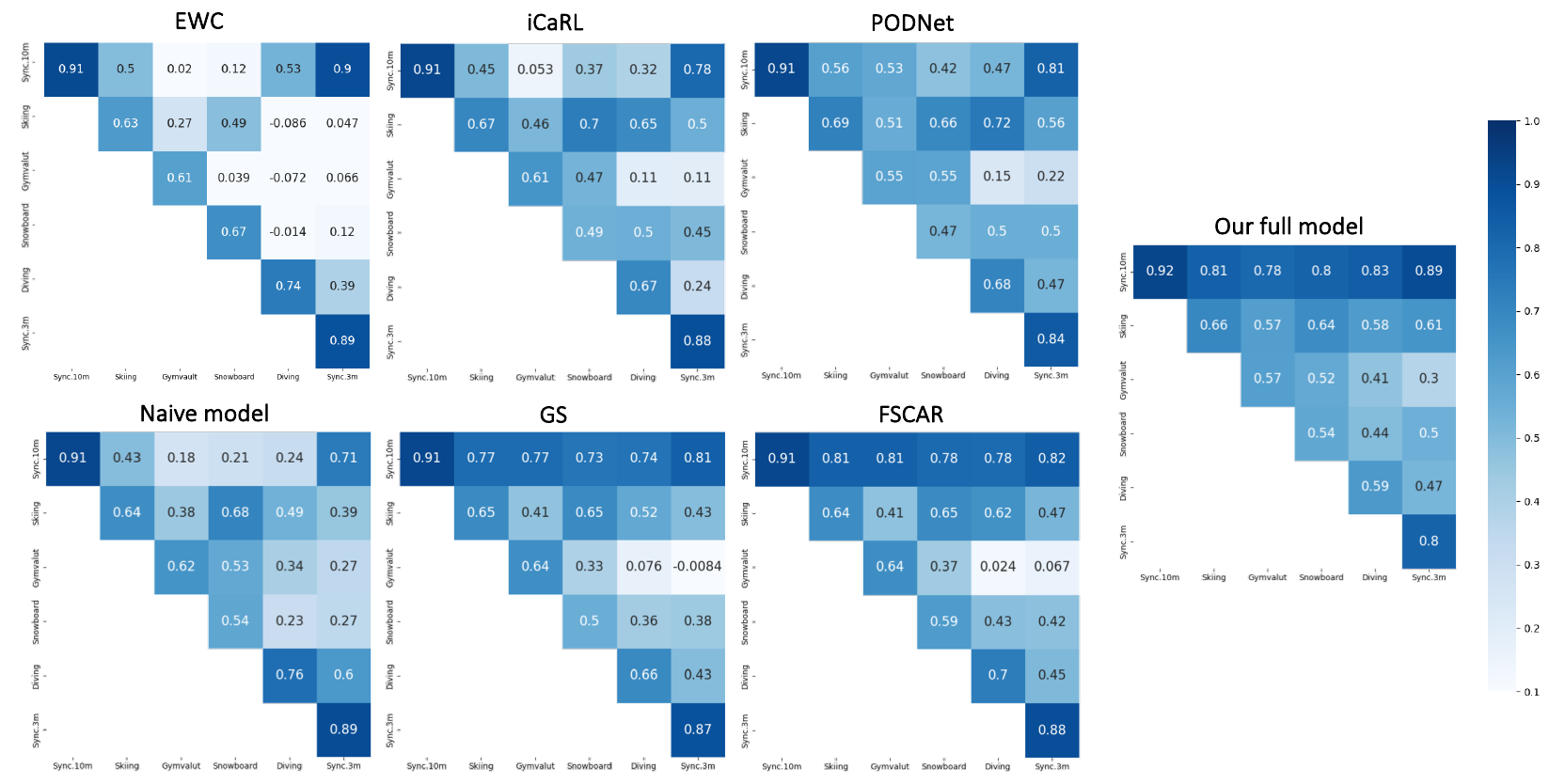}
   \caption{Visualizations of the performance matrix $P$ learned by EWC \cite{ewc}, iCaRL \cite{icarl}, PODNet \cite{podnet}, our naive model, our GS, our FSCAR and our full model on AQA-7 dataset. The \emph{i-row-j-column} element of $P$ denotes the Spearman’s rank correlation coefficient on the $i^{th}$ task after learning the $j^{th}$ task. \textbf{Darker color indicates better performance}. The task sequence is: Sync.10m, Skiing, Gym vault, Snowboard, Diving, Sync.3m.}
   \label{fig:vis1}
\end{figure*}

\vspace{0.13cm}
\subsection{Comparison results on the BEST and MTL-AQA datasets}
To study the versatility of our method, we conduct comparisons on two other popular AQA datasets (i.e., BEST \cite{best} and MTL-AQA \cite{mtl-aqa}) 
As shown in Table~\ref{tab:comparison_best} and Table~\ref{tab:comparison_mtl_aqa}, we have the same observation as experiments on the AQA-7 dataset: our method outperforms other CL baselines on all metrics.
Such results demonstrate that our method can adapt to the pairwise ranking scenario and works well in the scenario where continually learned tasks are pretty similar to the already learned ones.

\subsection{Intuitive Visualizations}
To further show the effectiveness of our model more intuitively, \ym{we conduct several visualizations.}

First, we visualize the performance matrix $P$, which has been introduced in Section~\ref{sec:dataset_metric}.
As shown in Figure~\ref{fig:vis1}, EWC \cite{ewc} suffers from serious forgetting. iCaRL \cite{icarl} mitigates the forgetting on the $2^{nd}$ and $4^{th}$ task, but preforms poorly on other tasks. PODNet \cite{podnet} outperforms iCaRL \cite{icarl} but still forgets a lot on the $2^{nd}$ task. Our approach achieves the best performance with the least forgetting.
Moreover, the model achieves better performance when gradually adding the proposed components.

\begin{figure*}[t]
  \centering
   \includegraphics[width=0.95\linewidth]{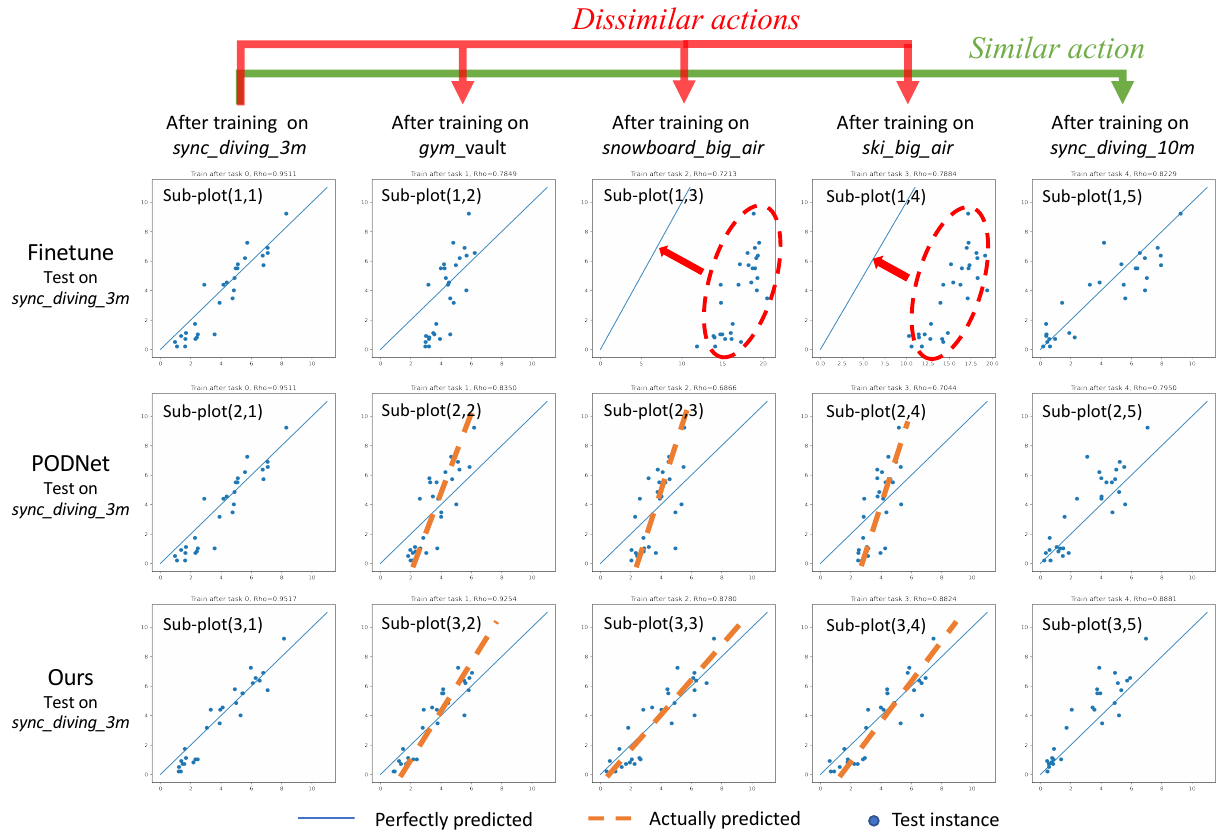}
   \caption{Visualizations of the assessment results on the first task during continual training obtained by Fintune, PODNet and Our full model. In each plot, more points close to the blue lines indicates better prediction of the ground-truth scores. The task sequence is: Sync.3m, Gym vault, Snowboard, Skiing, and Sync.10m. Therefore, for the first task, the $2^{nd}$, $3^{rd}$ and $4^{th}$ tasks are of dissimilar actions; the $5^{th}$ task is of similar action. The results show that our method can predict robust assessment results even after trained on dissimilar tasks.}
   % 图注表示纵坐标
   \label{fig:rho_evolution_plot}
\end{figure*}

Second, \ym{we further provide visualizations of the assessment results on the first task during continual training obtained by different methods (i.e., Finetune, PODNet \cite{podnet} and Ours) in Figure~\ref{fig:rho_evolution_plot}. 
When directly fine-tuning the model from one task to another one of dissimilar action, the model will predict scores far away from the ground-truth scores (i.e., Sub-plot(1-3) and Sub-plot(1-4)), and such problem would be alleviated after trained on tasks of similar actions (i.e., Sub-plot(1-5)). 
Although PODNet predicts scores much closer to the ground-truth labels, the model cannot predict promising scores after training on dissimilar tasks (i.e., Sub-plot(2-2), Sub-plot(2-3), and Sub-plot(2-4)).
Compared with fine-tuning and PODNet, our method achieves the best performance with the predicted scores closest to the ground-truth labels even after training the model on dissimilar tasks (i.e., Sub-plot(3-2), Sub-plot(3-3) and Sub-plot(3-4)). The visualization results further show that our method can achieve robust assessment performance even after trained on dissimilar tasks.
}

\section{Conclusions and Discussions}
We conduct a work on continual learning in AQA, which has never been explored before. We claim that the key to addressing Continual-AQA is sequentially learning a task-consistent score-discriminative feature distribution across all seen tasks. To address this, we try to mitigate forgetting in Continual-AQA in two aspects.
First, we propose a Feature-Score Correlation-Aware Rehearsal, which stores and reuses data from previous tasks with limited memory size, considering the relationship between latent features and action quality scores. 
Besides, to address the second question, an Action General-Specific Graph module is proposed to learn and decouple action-general and action-specific knowledge so that forgetting can be further mitigated.

The extensive experiments verified the effectiveness and versatility of the proposed approach. Although significant improvement has been achieved by our approach, Continual-AQA is still desirable to explore for the following reasons. 
First, the performance gap between our approach and the joint training upper bound is still explicit even when using a nontrivial memory size. 
Second, in this work, we just develop our solution based on the Directly Regression formulation of AQA (i.e., using a regressor to regress the action quality score directly). However, successful AQA models based on other formulations(e.g., Contrastive Regression \cite{gart}, and Distribution Prediction \cite{usdl}) are much easier to forget previously learned knowledge.

We believe the introduction of Continual-AQA and our exploration can provide a new perspective and solution for the practical application of AQA.

{\small
\bibliographystyle{IEEEtran}
\bibliography{egbib}
}

\begin{IEEEbiography}[{\includegraphics[width=1in,height=1.25in,clip,keepaspectratio]{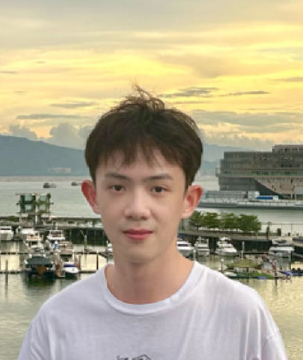}}]{Yuan-Ming Li} is received the B.S. degree in Software Engineering from University of Electronic Science and Technology of China in 2022. He is currently working toward the Ph.D. degree in Computer Science and Technology with the School of Computer Science and Engineering in Sun Yat-sen University. His research interests are computer vision and machine learning. He is currently focusing on the topic of action understanding.
\end{IEEEbiography}

\begin{IEEEbiography}[{\includegraphics[width=1in,height=1.25in,clip,keepaspectratio]{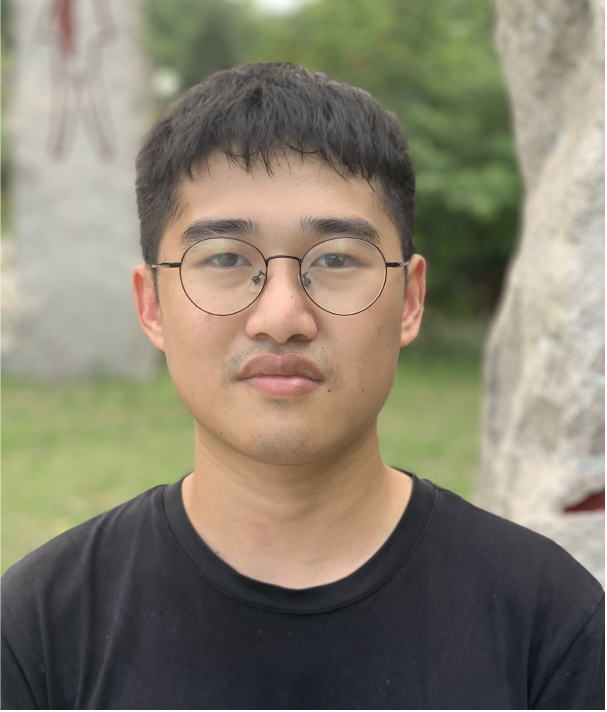}}]{Ling-An Zeng} is received the B.S. degree in Software Engineering from University of Electronic Science and Technology of China in 2019 and the M.S. degree in Computer Technology from Sun Yat-sen University in 2021. He is currently working toward the Ph.D. degree in Computer Science and Technology with the School of Artificial Intelligence in Sun Yat-sen University. His research interests are computer vision and machine learning. He is currently focusing on the topic of action understanding.
\end{IEEEbiography}

\begin{IEEEbiography}[{\includegraphics[width=1in,height=1.25in,clip,keepaspectratio]{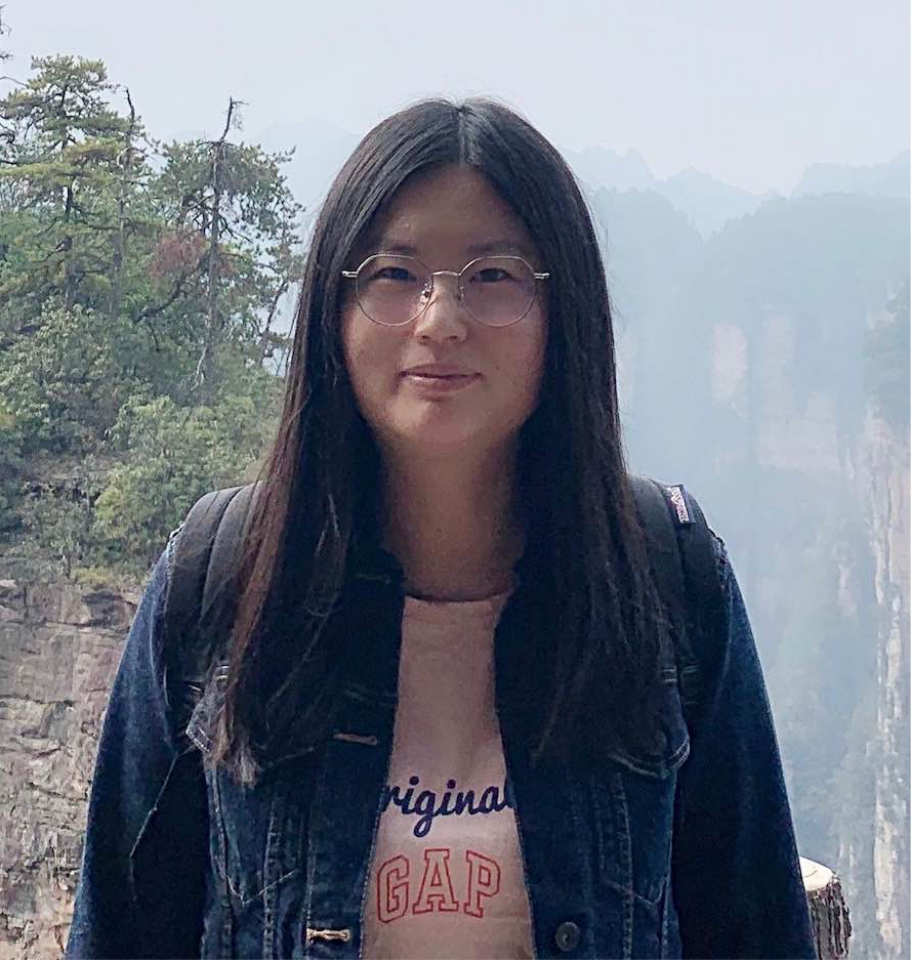}}]{Jing-Ke Meng} is an associate research fellow at the School of Computer Science and Engineering at Sun Yat-sen University, where she received her Ph.D. in Computer Science and Technology in 2020. Her research interests include computer vision and multi-modal learning.
\end{IEEEbiography}

\begin{IEEEbiography}[{\includegraphics[width=1in,height=1.25in,clip,keepaspectratio]{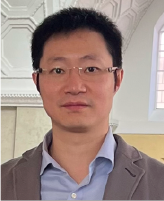}}]{Wei-Shi Zheng} is now a full Professor with Sun Yat-sen University. His research interests include person/object association and activity understanding, and the related weakly supervised/unsupervised and continuous learning machine learning algorithms. He has now published more than 200 papers, including more than 150 publications in main journals (TPAMI, IJCV, SIGGRAPH, TIP) and top conferences (ICCV, CVPR, ECCV, NeurIPS). He has ever served as area chairs of ICCV, CVPR, ECCV, BMVC, NeurIPS and etc. He is associate editors/on the editorial board of IEEE-TPAMI, Artificial Intelligence Journal, Pattern Recognition. He has ever joined Microsoft Research Asia Young Faculty Visiting Programme. He is a Cheung Kong Scholar Distinguished Professor, a recipient of the NSFC Excellent Young Scientists Fund, and a recipient of the Royal Society-Newton Advanced Fellowship of the United Kingdom.
\end{IEEEbiography}

\clearpage
\onecolumn

\noindent{\textbf{\large A.1 More Details about feature encoder with AGSG}}
\vspace{0.2cm}

We share the exact graph definition with JRG \cite{JRG, JRG_ASS}. Specifically, nodes represent latent representations of local patches, and edges represent the relationships between node pairs. As JRG, we get skeleton information in advance, and the patches are cropped around the human joints. Given local patch features from task $g$, we first calculate the AGSG $A^g$ as introduced in Eq. 7 in our manuscript. After that, local feature aggregations are performed. Notably, to aggregate local features across spatial and temporal spans, Spatial-AGG, Temporal-AGG, Spatial-ASG, and Temporal-ASG are introduced as the Spatial-JRG and Temporal-JRG in \cite{JRG_ASS}.
    
    Formally, given a video $v$, the extracted feature can be written as: 
    \begin{equation}
        E_g(v) = Concatenate(v_w,v_p,H^g,D_s^g,D^g_r),
    \end{equation}
    where $v_w\in \mathbb{R}^{T\times D}$ and $v_p\in \mathbb{R}^{T\times J\times D}$ indicate the whole-scene features and the local patch features extracted by pre-trained I3D backbone. Here $T$, $J$, and $D$ denote the number of temporal steps, the number of joints, and the dimension of features, respectively. Following \cite{JRG_ASS}, we use $H^g$, $D_s^g$, and $D^g_r$ to denote the motion commonality features, spatial difference features, and temporal difference features.

    Specifically, given local patch features $v_p$, the motion commonality features $H^g$ is calculated as:
    \begin{equation}
        \begin{split}
        H^g & = \alpha * G_s * v_p + (1-\alpha) * S_s^g * v_p \\
        & = (\alpha * G_s + (1-\alpha) * S_s^g) * v_p \\
        & = A^g_s * v_p,
        \end{split}
    \end{equation}
    where $G_s$ and $S_s^g$ indicate Spatial-AGG, and Spatial-ASG for task $g$.
    Additionally, the spatial difference features and temporal difference features are calculated as follows:
    \begin{equation}
        \begin{split}
        D^{g,t}_s(i) & = \sum_j \alpha\cdot (G_s(i,j)\cdot (v_p^t(i)-v_p^{t-1}(j)))\cdot w_j + (1-\alpha)\cdot (S_s^g(i,j)\cdot (v_p^t(i)-v_p^{t-1}(j)))\cdot w_j \\
        & = \sum_j (\alpha\cdot G_s(i,j) + (1-\alpha)\cdot S_s^g(i,j))\cdot (v_p^t(i)-v_p^{t-1}(j))\cdot w_j \\
        & = \sum_j A^g_s(i,j) \cdot v_p^t(i)\cdot w_j, 
        \end{split}
    \end{equation}
    \begin{equation}
        \begin{split}
        D^{g,t}_r(i) & = \sum_j \alpha\cdot (G_r(i,j)\cdot (v_p^t(i)-v_p^{t-1}(j)))\cdot w_j + (1-\alpha)\cdot (S_r^g(i,j)\cdot (v_p^t(i)-v_p^{t-1}(j)))\cdot w_j \\
        & = \sum_j (\alpha\cdot G_r(i,j) + (1-\alpha)\cdot S_r^g(i,j))\cdot (v_p^t(i)-v_p^{t-1}(j))\cdot w_j \\
        & = \sum_j A^g_r(i,j) \cdot v_p^t(i)\cdot w_j, 
        \end{split}
    \end{equation}
    where $1\leq i,j \leq J$. $G_r$ and $S_r^g$ indicate Temporal-AGG, and Temporal-ASG for task $g$. 
    $w_j$ is the weight aggregating the spatial/temporal difference in the joint neighborhood.
    $D^{g,t}_s(i)\in \mathbb{R}^D$ and $D^{g,t}_r(i)\in \mathbb{R}^D$ are the spatial difference features of joint $i$ at time step $t$. We further obtain $D^g_s\in \mathbb{R}^{T\times J\times D}$ and $D^g_r\in \mathbb{R}^{T\times J\times D}$ via concatenating all $D^{g,t}_s(i)$ and $D^{g,t}_r(i)$ by joint and temporal dimension. After all, we reshape $v_w,v_p, H^g, D_s^g, D^g_r$ to the same size and concatenate the output feature map. 
    
    Note that before using the regressor to predict the score or score difference, average pooling will be adopted on the output feature map, and a linear layer followed by a ReLU function will be utilized to reshape the feature size further.
% \section{Algorithm}
% \newpage
\vspace{0.5cm}

\noindent{\textbf{\large A.2 Anchor-based Helper Selection}}
\vspace{0.2cm}

Here we provide more details about the anchor-based helper selection strategy.

We first note all the stored data sorted by the score labels as $\psi=\{(v_1,s_1),...,(v_M,s_M)\}$, where $s_i < s_{i+1}$. Suppose now we are going to select a helper set $H=\{(v_{hlp,1},s_{hlp,1}),...,(v_{hlp,K},s_{hlp,K})\}$ from $\psi$ to augment the extracted feature and score of an anchor data $(v_i,s_i)$. 
We hope that the $H$ can contain both higher and lower score helpers than the anchor. At the same time, the ratio of helpers with higher scores to helpers with lower scores in $H$ should match the true data distribution. To this end, we separately sample lower-score helpers $H_{low}$ from $\{(v_1,s_1),...,(v_{i-1},s_{i-1})\}$ and higher-score helpers from $H_{high}$ from $\{(v_i,s_i),...,(v_{M},s_{M})\}$ with constrains that $|H_{low}|/|H_{high}|\approx (i-1)/(M-i)$, and $|H_{low}|+|H_{high}|=K$. Since such a helper-selecting strategy is based on the anchor data, we note it as \textbf{\emph{Anchor-based}} to distinguish it from the \textbf{\emph{Random}} strategy.
\vspace{0.5cm}

\noindent{\textbf{\large A.3 Training Algorithm}}
\vspace{0.2cm}

Algorithm~\ref{al:training} illustrates the details about how to train our model continually.
\begin{algorithm*}[hb]
    \caption{Training our model on a current task $t$ ($t>1$)}
    \label{al:training}
    \begin{algorithmic}[1]
        \Require                 %输入条件
        Previous feature extractor $E^{t-1}$; Previous score regressor $R^{t-1}_s$; Difference regressor $R^{t-1}_d$; Action-Specific Graph $S^t$ for task $t$; Current data $D^t$; Saved previous data $\psi=\{\psi^1,...,\psi^{t-1}\}$; Hyper-parameters $\lambda_{FD}, \lambda_{Diff}, M$.
        \Ensure {$E^{t}, R^{t}_s, R^{t}_d, \psi$}       %输出
        % \BlankLine 
        \vspace{0.3cm}
        \STATE $\text{Initialization: } E^{t}, R^{t}_s, R^{t}_d \gets E^{t-1}, R^{t-1}_s, R^{t-1}_d$
        \vspace{0.3cm}
        \STATE $E^{t} \gets E^{t} \cup S^t$
        \vspace{0.3cm}
        \WHILE{$\textnormal{until required iterations}$}
            % \STAET $D^t_{tr} \gets \text{Sample a mini-batch from }D_t$  \\
            % \vspace{0.3cm}
            \STATE $D^t_{tr} \gets \textnormal{Sample a mini-batch from } D_t$ 
            \vspace{0.3cm}
            \STATE $\mathcal{L}_{Diff},\mathcal{L}_{FD},\mathcal{L}_{AQA},N_{tr} \gets 0$ 
            \vspace{0.3cm}
            \FOR {$(v^t_{cur,i}, s_{cur,i}) \text{ in } D^t_{tr}$}
                % \vspace{0.3cm}
                \STATE $(v^\tau_{pre,j}, s_{pre,j}) \gets \text{RandomSample}(\psi)$               \Comment{\emph{Randomly sample previous data. $\tau$ indicates the task id}}
                \vspace{0.3cm}
                \STATE $\{(v^{\epsilon_k}_{hlp,k}, s_{hlp,k})\}^K_{k=1} \gets \text{RandomSample}(\psi)$     \Comment{\emph{Randomly sample K augmentation helpers}}
                \vspace{0.3cm}
                \STATE $\{f^{t,n}_{cur,i}\}^{t}_{n=1} \gets \{E^t_n(v^t_{cur,i})\}^{t}_{n=1}$  \Comment{\emph{Fed the current data into $E^{t}$}}
                \vspace{0.3cm}
                \STATE $\{f^{t-1,n}_{cur,i}\}^{t-1}_{n=1} \gets \{E^{t-1}_n(v^t_{cur,i})\}^{t-1}_{n=1}$  \Comment{\emph{Fed the current data into $E^{t-1}$}}
                \vspace{0.3cm}
                \STATE $\{f^{t,n}_{pre,j}\}^{t-1}_{n=1} \gets \{E^t_n(v^\tau_{pre,j})\}^{t-1}_{n=1}$   \Comment{\emph{Fed the previous data into $E^{t}$}}
                \vspace{0.3cm}
                \STATE $\{f^{t-1,n}_{pre,j}\}^{t-1}_{n=1} \gets \{E_n^{t-1}(v^\tau_{pre,j})\}^{t-1}_{n=1}$   \Comment{\emph{Fed the previous data into $E^{t-1}$}}
                \vspace{0.3cm}
                \STATE $\{f^{t-1,{\epsilon_k}}_{hlp,k}\}^K_{k=1} \gets \{E^{t-1}_{\epsilon_k}(v^{\epsilon_k}_{hlp,k})\}^K_{k=1}$       \Comment{\emph{Fed the augmentation helpers into $E^{t-1}$}}
                \vspace{0.3cm}
                \STATE $(\tilde{f}^{t-1}_{pre,j},\tilde{s}_{pre,j}) \gets \text{FS-Aug}( (f^{t-1}_{pre,j},s_{pre,j}), \{(f^{t-1}_{hlp,k}, s_{hlp,k})\}^K_{k=1})$ \Comment{\emph{Feature-Score co-Augmentation}}
                \vspace{0.3cm}
                \STATE $\hat{s}_{pre,j},\hat{s}_{cur,i} \gets R^t_s(f^{t,\tau}_{pre,j}), R^t_s(f^{t,t}_{cur,i})$    \Comment{\emph{Computer predicted scores}}
                \vspace{0.3cm}
                \STATE $d \gets \tilde{s}_{pre,j} - s_{pre,j}$   \Comment{\emph{Computer GT score difference via Eq.~5}}
                \vspace{0.3cm}
                \STATE $\mathcal{L}_{Diff} \gets \mathcal{L}_{Diff}+(R^t_d(\tilde{f}^{t-1}_{pre,j}\copyright f_{pre,j}^{t})-d)^2$ \Comment{\emph{Compute $\mathcal{L}_{Diff}$ via Eq.~6}}
                \vspace{0.3cm}
                \STATE $\mathcal{L}_{FD} \gets \mathcal{L}_{FD}+\sum^{t-1}_{n=1}[||f^{t,n}_{cur,i}-f^{t-1,n}_{cur,i}||^2+||f^{t,n}_{pre,j}-f^{t-1,n}_{pre,j}||^2]/2$ \Comment{\emph{Compute $\mathcal{L}_{FD}$ via Eq.~8}}
                \vspace{0.3cm}
                \STATE $\mathcal{L}_{AQA} \gets \mathcal{L}_{AQA}+ ((\hat{s}_{pre,j}-s_{pre,j})^2 + (\hat{s}_{cur,i}-s_{cur,i})^2)/{2} $ \Comment{\emph{Compute $\mathcal{L}_{AQA}$ via MSE loss}}
                \vspace{0.3cm}
                \STATE $N_{tr} \gets N_{tr}+1$  
            % \vspace{0.3cm}
            \EndFor
            % \vspace{0.3cm}
            \STATE $\mathcal{L}_{Diff},\mathcal{L}_{FD},\mathcal{L}_{AQA} \gets {\mathcal{L}_{Diff}}/{N_{tr}}, {\mathcal{L}_{FD}}/{N_{tr}}, {\mathcal{L}_{AQA}}/{N_{tr}}$ 
            \vspace{0.3cm}
            \STATE $\mathcal{L} \gets \mathcal{L}_{AQA} + \lambda_{FD} \mathcal{L}_{FD} + \lambda_{Diff} \mathcal{L}_{Diff}$
            \vspace{0.3cm}
            \STATE $\text{Update } E^{t}, R^{t}_s, R^{t}_d \text{ by back-propagation}$
        \vspace{0.3cm}
        \EndWhile
        \vspace{0.3cm}
        \STATE $\psi \gets \text{GS-Reconstruction}(\psi, M, D^t)$ \Comment{\emph{Reconstruct $\psi$ via GS}}
        % \RETURN 
    \end{algorithmic}
    \end{algorithm*}

\clearpage

\end{document}